%% file: arxiv_v2.tex
\documentclass{article}
\usepackage{arxiv}

\usepackage[T1]{fontenc}
\usepackage{url}
\usepackage[colorlinks,citecolor=blue,urlcolor=blue,backref=page,backref=page,linkcolor=blue]{hyperref}
\usepackage{booktabs}
\usepackage{amsfonts}
\usepackage{nicefrac}
\usepackage{microtype}
\usepackage{graphicx}
\usepackage{natbib}
\usepackage{siunitx}
\usepackage{appendix}
\usepackage{float}
\usepackage{amsthm}
\usepackage{amsmath}
\usepackage{amsfonts}
\usepackage{amssymb}
\usepackage{subcaption}
\usepackage{cleveref}
\usepackage[export]{adjustbox}

\title{Amortized Bayesian Multilevel Models}

\author{
Daniel Habermann \\
Department of Statistics \\
TU Dortmund University \\ Germany
\And
Marvin Schmitt \\
Cluster of Excellence SimTech \\
University of Stuttgart \\  Germany
\And
Lars Kühmichel \\
Department of Statistics \\
TU Dortmund University \\ Germany
\And
Andreas Bulling \\
Institute for Visualisation and Interactive Systems \\
University of Stuttgart \\ Germany
\And
Stefan T. Radev \\
Department of Cognitive Science \\
Rensselaer Polytechnic Institute \\ USA
\And
Paul-Christian Bürkner \\
Department of Statistics \\
TU Dortmund University \\ Germany
}

\renewcommand{\headeright}{}
\renewcommand{\undertitle}{}

\hypersetup{
pdftitle={Amortized Bayesian Multilevel Models},
pdfauthor={Daniel Habermann, Marvin Schmitt, Lars Kühmichel, Andreas Bulling, Stefan T. Radev, and Paul-Christian Bürkner},
pdfkeywords={Bayesian Models, Amortized Inference, Multilevel Models},
}

\date{}

\input{_math}

\begin{document}
\maketitle

\begin{abstract}
Multilevel models (MLMs) are a central building block of the Bayesian workflow. 
They enable joint, interpretable modeling of data across hierarchical levels and provide a fully probabilistic quantification of uncertainty. 
Despite their well-recognized advantages, MLMs pose significant computational challenges, often rendering their estimation and evaluation intractable within reasonable time constraints. 
Recent advances in simulation-based inference offer promising solutions for addressing complex probabilistic models using deep generative networks.
However, the utility and reliability of deep learning methods for estimating Bayesian MLMs remains largely unexplored, especially when compared with gold-standard samplers.
To this end, we explore a family of neural network architectures that leverage the probabilistic factorization of multilevel models to facilitate efficient neural network training and subsequent near-instant posterior inference on unseen datasets. 
We test our method on several real-world case studies and provide comprehensive comparisons to Stan's gold standard sampler, where possible. 
Finally, we provide an open-source implementation of our methods to stimulate further research in the nascent field of amortized Bayesian inference.
\end{abstract}


\maketitle

\section{Introduction}

Accurate inference and reliable uncertainty quantification in reasonable time is a frontier of today's statistical research \citep{Cranmer2020}.
One major difficulty arising in almost all experimental and observational data is the presence of complex dependency structures, such as natural groupings (e.g., data gathered in different countries) or repeated measurements of the same observational units over time \citep[e.g., patients in clinical trials or students in educational studies;][]{Gelman2006, Fitzmaurice2011}.
To reflect these dependency structures, multilevel models (MLMs), also referred to as latent variable, hierarchical, random, or mixed effects models, have become an integral part of modern Bayesian statistics \citep[e.g.,][]{goldstein2011multilevel, Gelman2013, mcglothlin2018bayesian, finch2019multilevel, yao2022bayesian}.
Despite the wide success of Bayesian MLMs across the quantitative sciences, a major challenge is their limited scalability, as estimating the full posterior distribution of all parameters of interest can be very costly \citep{Gelman2013, betancourt2017conceptual}.
For models where the likelihood function is analytically tractable and differentiable with respect to model parameters, Markov chain Monte Carlo (MCMC) sampling algorithms as implemented in probabilistic programming languages like Stan \citep{stan_2024} are the current gold standard for generating accurate draws from the posterior distribution.
Recently, there has been a growing body of research that is concerned with using pairs of samples from the prior distribution and simulated datasets to train neural networks that can approximate the posterior for unseen data \citep[e.g.,][]{papamakarios_fast_2016, wildberger2024flow, kobyzev2020normalizing}. The prospect of these neural posterior estimation (NPE) methods is that they can enable \textit{amortized} inference \citep{gershman2014amortized}, which refers to the property that posterior inference for new data is almost instant after training of the neural networks has been completed \citep{Radev2020a, tejero-cantero2020sbi}.

Expanding neural density estimation to MLMs is challenging because the dimension of the posterior depends on the number of groups and may vary over datasets. Such settings require specialized neural architectures to lift the requirement of fixed-length inputs.
In MLMs, amortized inference with varying group sizes is complicated by the fact that estimates of local parameters for one group depend not only on that specific group but also on all other groups in the dataset.
Additionally, the number of parameters increases linearly with the number of groups, rendering network training inefficient for datasets with a large number of groups.

In important pioneering work, initial progress has been made in addressing these challenges \citep{rodrigues2021hnpe, Arruda2023, heinrich_hierarchical_2023}, but they have so far considered only a narrow range of MLMs and have performed only limited validation against state-of-the-art samplers for Bayesian inference. 
In particular, existing methods do not consider (1) simultaneous amortization over both the number of groups and the number of observations per group, (2) amortization in models with covariates, (3) simultaneous joint estimation with non-IID response variables, (4) highly correlated group-level parameters, (5) additional (non-hierarchical) shared parameters, as well as (6) systematic calibration and shrinkage analysis to verify inference correctness at each level. Our proposed framework addresses these challenges by making the following contributions:
\begin{itemize}
    \item We develop neural network architectures that utilize the probabilistic factorization of the likelihood of multilevel models to facilitate efficient neural network training and subsequent near-instant amortized posterior inference.
    \item We test our method on a wide range of real-world case studies which together cover all of the above mentioned challenges. We also provide comparisons to Stan as a gold-standard method for posterior inference whenever possible.
    \item We provide an efficient and user-friendly implementation of our algorithm in the \texttt{BayesFlow} Python library \citep{radev2023joss}. In this way, users can benefit from amortized inference for otherwise intractable models and generate fast and accurate draws from the posterior distribution.
\end{itemize}

We start by briefly discussing the limitations of MCMC-based sampling methods and current research avenues (\autoref{sec:mcmc-challenges}). We then follow by introducing neural posterior estimation, highlighting possible advantages over traditional MCMC-based methods as well as current shortcomings (\autoref{sec:npe}). Our novel contributions start from \autoref{sec:ml-npe} onward, where we extend NPE to multilevel models and address challenges specific to hierarchical data. For readers without a background in machine learning, we provide an overview of the current landscape and introduction to ``simulation-based inference'', ``amortized Bayesian inference'', and ``neural density estimation'' in \autoref{sec:supp:terminology}.
\autoref{sec:experiments} evaluates multilevel NPE through a variety of experiments, focusing on model classes that were previously not amenable to neural posterior estimation.

\subsection{Challenges of Posterior Inference in Bayesian Multilevel Models} \label{sec:mcmc-challenges}

Considerable effort has been made to improve the sampling speed of MCMC algorithms, especially for high-dimensional models with rich structure, such as MLMs. This includes step-size adjustments or reparameterizations to work-around or improve the posterior geometry \citep[e.g.,][]{hoffman_neutralizing_2019, modi_delayed_2023, bironlattes2024, girolami2011}, enabling within-chain parallelization via parallel evaluation of the likelihood \citep{stan_2024, Lee2010}, as well as the development of specialized algorithms that entail faster and more reliable adaptation phases, such that multiple short chains can be run in parallel \citep[e.g.,][]{Zhang2022, Margossian2024, rosenthal2000parallel, lao2020tfp, hoffman2021}.
However, despite these advancements, sampling methods based on MCMC are still too slow for many multilevel settings. Even for models and datasets of moderate size, posterior inference can take days or even weeks, creating a large gap between the models that researchers aim to compute (which may include more complex structures) and the simplified models that are computationally feasible with current methods. Expanding the space of models that can be fit in a reasonable time frame is therefore an important requirement for the broad applicability of Bayesian MLMs. 

To address these challenges, approximate inference methods such as variational inference \citep[e.g.,][]{hoffman2015, agrawal2021} or integrated Laplace approximation \citep[INLA, e.g.,][]{rue2009, rue2017bayesian, margossian2020hamiltonian} have been developed. Such algorithms trade off speed against posterior accuracy or modeling flexibility \citep[e.g.,][]{yao_yes_2018, mackay2003information, blei2017variational, margossian2023shrinkage}. In cases where model parameters and their uncertainty are of interest, and not only model predictions, the loss of posterior accuracy is particularly problematic, as it may invalidate the conclusions drawn from the model's inference – even if the model itself is reasonable.

In addition to sampling speed, there are two fundamental issues with posterior inference for Bayesian MLMs in most applications of scientific interest:

\begin{enumerate}
    \item The need to rerun computationally expensive sampling algorithms when new data becomes available or when the quality of model fits needs to be assessed.
    \item Generating accurate draws from the posterior when the likelihood is intractable or non-differentiable.
\end{enumerate}

Issue (1) is a crucial limiting factor in cases where data are arriving in real-time or many datasets need to be evaluated. In addition, essential steps in the modern Bayesian workflow \citep{Gelman2020} like cross-validation \citep[e.g.,][]{Vehtari2017, merkle_bayesian_2019} or simulation-based calibration \citep[SBC, e.g.,][]{Cook2006, Talts2018, modrak2023simulation} require many model refits on subsets of the data, rendering careful model testing a time-intensive task for all but the simplest models. Issue (2) is also becoming increasingly common, as many scientific domains nowadays employ models whose output is the result of a complex simulation program \citep{Cranmer2020}. Consequently, closed-form solutions of the likelihood are often unavailable.

Both of these issues have led to the development of novel algorithms that do not rely on MCMC sampling.
For models without a tractable likelihood, simulation-based inference methods such as approximate Bayesian computation \citep[ABC,][e.g.,]{Sisson2018} can sample from the posterior without computing the likelihood directly. The main drawback of ABC compared to neural network based approaches is that it heavily suffers from the curse of dimensionality \citep[e.g.,][]{Barber2015, beneventano2020high} as the number of parameters increases. This is of particular concern for MLMs as their number of model parameters scales linearly with the number of groups in the data. While ABC algorithms for MLMs have been developed \citep{sisson2011likelihood}, they have not been able to fully overcome this major drawback of ABC. Other approaches to Bayesian inference for MLMs with intractable likelihoods include mean-field variational algorithms \citep[e.g.,][]{Tran2017, Roeder2019}, but this class of algorithms tends to come with the aforementioned loss of posterior accuracy \citep[e.g.,][]{yao_yes_2018}.

\section{Methods} \label{sec:methods}

\begin{figure}[ht!]
    \centering
    \begin{minipage}{0.32\textwidth}
        \includegraphics[width=\textwidth]{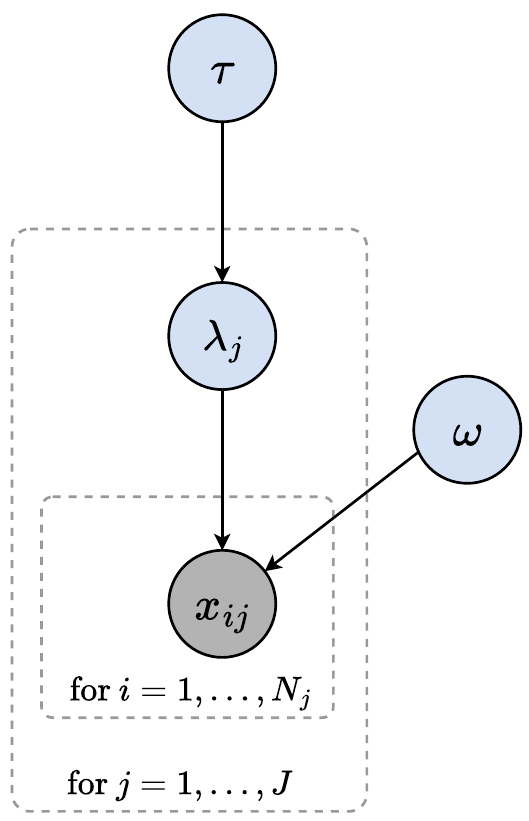}
    \end{minipage}
    \hfill
    \begin{minipage}{0.55\textwidth}
        \hspace{-0.5cm}
        \includegraphics[width=1.1\textwidth]{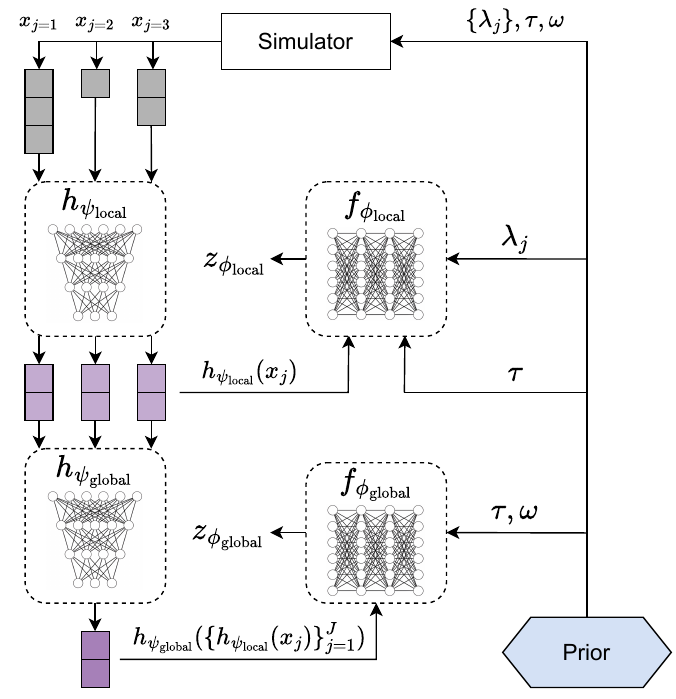}
    \end{minipage}
        \caption{\textbf{Left}: Directed graph of a two-level Bayesian model with hyperparameters $\tau$, local parameters $\lambda_j$, shared parameters $\omega$ and observed data $x_{ij}$. Gray-shaded circles indicate observed values, whereas non-shaded circles indicate latent variables. Dashed boxes represent exchangeability. 
        \textbf{Right}: Summary of the multilevel NPE architecture. 
        Simulated datasets with varying group sizes (grey stacked boxes) are transformed into fixed-length representations by a local summary network $h_{\psi_{\text{local}}}$ (purple squares). A global summary network $h_{\psi_{\text{global}}}$ then converts these local summaries into a fixed-length dataset representation. The local inference network $f_{\phi_{\text{local}}}$ maps prior draws of $\lambda_j$ to a base distribution $z_{\phi_\text{local}}$, conditioned on local summaries and $\tau$. The global inference network $f_{\phi_{\text{global}}}$ similarly maps prior draws of $\tau, \omega$ to a base distribution $z_{\phi_{\text{global}}}$, conditioned on the global dataset summary.}
        \label{fig:twolevel-graph-and-architecture}
\end{figure}

\subsection{Notation and Definitions} \label{sec:notation}

We denote the joint distribution over both the parameters $\theta$ and observed data $x$ of a parametric Bayesian model as $p(\theta, x)$. 
We assume that $\theta \in \mathbb{R}^D$ represents all unobservable (latent) quantities, whereas $x \in \mathcal{X}$ denotes all observable quantities.
The joint distribution $p(\theta, x)$ factorizes into a data model $p(x \given \theta)$ and a prior $p(\theta)$, defining a ``generative model'' that can be translated into a computer program and sampled from according to a simple ancestral sampling scheme:
\begin{equation}
    \begin{split}
        \theta &\sim p(\theta)\\
        x &\sim p(x \given \theta)
    \end{split}
\end{equation}
Multilevel models can be expressed as a special case of this general notation, where the prior $p(\theta)$ is factorized in an application-dependent manner \citep{robert2007bayesian}. For example, in a two-level model with exchangeable groups $j$ and exchangeable observations within groups $i$, there are three qualitatively different kinds of parameters: (1) \textit{Local parameters} $\lambda_j$ indexed by $j \in \{1, \dots, J\}$ that are specific to each group $j$, such as separate intercepts or slopes per group in a regression model; (2) \textit{hyperparameters} $\tau$, such as means and standard deviations for the local parameters, or correlations of local parameters in the same group; and (3) \textit{shared parameters} $\omega$ with a direct effect on the likelihood function of all groups, without any hierarchical component (see \autoref{fig:twolevel-graph-and-architecture} for a diagram of the dependency structure, left panel). We use the term \textit{global parameters} to refer to the hyperparameters $\tau$ and shared parameters $\omega$ jointly. 

Thus, in a two-level setting, the full set of parameters is defined as $\theta := \{\{\lambda_j\}_{j=1}^{J}, \tau, \omega\}$. Consequently, the prior $p(\theta)$ factorizes as:
\begin{equation}
    p(\theta) = p(\tau, \omega, \{\mathbf{\lambda}_j\}_{j=1}^{J}) = p(\tau)  \; p(\omega) \prod_{j=1}^{J} p(\lambda_j \given \tau)
\end{equation}
Let $x_{ij}$ denote the observation with index $i \in 1,\dots,N_j$ within group $j$. The generative model can then be described as (see \autoref{fig:twolevel-graph-and-architecture}):
\begin{equation}
    \begin{aligned}
        \tau, \omega &\sim p(\tau, \omega)\\
        \lambda_j \given \tau &\sim p(\lambda \given \tau) && \textbf{for}\,\,j = 1,\dots,J\\
        x_{ij} &\sim p(x \given \lambda_j, \omega) &&\textbf{for}\,\,i = 1,\dots,N_j
    \end{aligned}
\end{equation}
To fit these multilevel models in practice, we first introduce the concept of neural posterior estimation in \autoref{sec:npe}, before expanding it to multilevel NPE in \autoref{sec:ml-npe}.

\subsection{Neural Posterior Estimation} \label{sec:npe}

Recently, algorithms based on neural posterior estimation \citep[NPE, e.g.,][]{greenberg_automatic_2019, Radev2020, gonccalves2020training} have emerged as a promising alternative to classical MCMC.
In NPE, generating draws from the posterior distribution of a Bayesian model is achieved by transforming a random input vector, e.g. sampled from a unit Gaussian, into draws from the target posterior distribution using specialized neural architectures. 
Below, we focus on NPE with normalizing flows \citep[e.g.,][]{kobyzev2020normalizing} as a concrete example, although other generative neural architectures can be used, such as diffusion models \citep[e.g.,][]{sharrock2022diffusionsbi}, consistency models \citep[e.g.,][]{schmitt_consistency_2024}, or flow matching \citep[e.g.,][]{wildberger2024flow}.

A normalizing flow is an invertible transformation between the target (posterior) distribution and a tractable base distribution, such as a unit Gaussian. Let $z \in \mathbb{R}^D$ be a $D$-dimensional random variable with tractable density, e.g. $p_0(z) = \text{Normal}(z \given 0, \mathbb{I})$, and let $\theta \in \mathbb{R}^D$ be a random variable with unknown probability density equivalent to the posterior $p(\theta \given x)$ over model parameters $\theta$ given observed data $x$. Let $f : \mathbb{R}^D \rightarrow \mathbb{R}^D$ be an invertible and differentiable function such that $z = f(\theta)$ and $\theta=f^{-1}(z)$.
In practice, this function is usually parameterized in terms of a conditional invertible neural network $f_{\phi}$ with learnable network weights $\phi$.

The \textit{inference network} $f_{\phi}$ maps model parameters $\theta$ to latent Gaussian variates $z$ using conditional information from the observed dataset $x$. Conversely, the inverse pass through the network maps $z$ to $\theta$ using the same conditional information. 
Training data $\{(\theta^{(m)},x^{(m)})\}^{M}_{m=1}$ for the neural network are generated by first sampling a single realization of model parameters $\theta^{(m)}$ from the prior $p(\theta)$ and then simulating a synthetic dataset $x^{(m)}$ of a specified size via the generative process (i.e., $x^{(m)} \sim p(x \given\ \theta^{(m)})$.
Because $f_\phi$ is differentiable, the density of the approximate posterior $q_\phi(\theta \given x)$ can be computed via the change-of-variables formula:
\begin{equation}
    q_{\phi}(\theta \given x) = p_0\big(f_{\phi}(\theta; x)\big)\left\lvert\det\dfrac{\partial f_{\phi}(\theta; x)}{\partial\theta}\right\rvert,
\end{equation}
where $f_\phi(\theta; x)$ denotes a forward pass of parameters $\theta$ through the neural network $f_\phi$, conditional on a dataset $x$ and $p_0(f_{\phi}(\theta;x))$ is the tractable density of the base distribution, e.g. with $p_0(z) = \text{Normal}(z \given 0, \mathbb{I})$. 
Training of the network weights $\phi$ is then achieved by minimizing the maximum likelihood loss:
\begin{equation}
    \label{eq:npe-simulation-based-loss}
    \mathcal{L}_{\text{NPE}}(\phi) = -\frac{1}{M}\sum_{m=1}^{M}\log q_\phi(\theta^{(m)} \given x^{(m)})
\end{equation}
Minimizing this loss during training for sufficiently large simulation budget $M$ results in minimizing the expected forward Kullback-Leibler (KL) divergence between the approximate posterior $q_\phi(\theta \given x)$ and the true posterior $p(\theta \given x)$:

\begin{equation}\label{eq:npe-loss}
\begin{aligned}
    \widehat{\phi} &= \argmin_{\phi} \mathbb{E}_{(\theta, x)\sim p(\theta, x)}\big[\mathrm{KL}\big( 
    p(\theta \given x)\,\big|\big|\,
    q_{\phi}(\theta \given x)\big)\big]
    \\ &=\argmin_{\phi} \mathbb{E}_{(\theta, x)\sim p(\theta, x)}\big[-\log q_{\phi}(\theta \given x)\big]
\end{aligned}
\end{equation}

One particular advantage of NPE is that, after an initial training phase, it enables near-instantaneous posterior sampling for unseen datasets without any further training required \citep[e.g.,][]{Radev2020, gonccalves2020training}. This property is called \textit{amortization} \citep{gershman2014amortized}, as the initial computational cost during training is later offset by allowing posterior inference for new datasets typically within a fraction of a second.
Amortization is a direct consequence of the fact that a single (global) set of neural network parameters $\phi$ is used to approximate the posterior distribution of many datasets in the support of $p(\theta, x)$.

Typically, the goal is to allow posterior inference for datasets with varying sample size $N \in \mathbb{N}$, which requires that the inference network is also trained with datasets of varying sample size. As the inference network can only handle fixed-length inputs, a second neural network, called \textit{summary network} is often used in practice \citep[e.g.,][]{Radev2020}. This summary network converts datasets with varying sample size into fixed-length representations, which are then used as conditions for the inference network. The summary network can also be beneficial for other reasons: If the data are high-dimensional, it may improve computational efficiency by using a lower-dimensional representation that is sufficient to inform posterior inference. Second, special neural architectures can be used to reflect the structure of the data. For example, permutation invariant networks can be used for exchangeable data to ensure that posterior inference does not depend on the order of individual data points. 

Throughout this work, we denote summary networks by $h_\psi$, where $\psi$ refers to the learnable network weights. 
The summary network $h_\psi$ and inference network $f_\phi$ are optimized in one joint \textit{end-to-end} optimization objective. Expanding \autoref{eq:npe-loss} with a summary network yields:
\begin{equation}
    \label{eq:npe-loss-joint}
    \widehat{\psi}, \widehat{\phi} = \argmin_{\psi, \phi} \mathbb{E}_{(\theta, x)\sim p(\theta, x)}\Big[-\log q_{\phi}\big(\theta \given h_{\psi}(x)\big)\Big],
\end{equation}
which can also be trained via a simulation-based loss as shown in \autoref{eq:npe-simulation-based-loss}.
The learned summary statistics are then \textit{approximately sufficient} \citep{chen2020neural, Radev2020} for posterior inference. The term approximately sufficient describes the fact that the learned summary statistics $h_\psi(x)$ are generally not sufficient to reconstruct the dataset $x$ itself, but are sufficient in the sense that they do not alter the posterior distribution when swapped with the data: $p(\theta \given x) \approx p(\theta \given h_\psi(x))$.

\subsection{Multilevel Neural Posterior Estimation} \label{sec:ml-npe}

Standard NPE is not directly suitable for estimating multilevel models without further modifications. This limitation arises from the hierarchical structure of multilevel models, where parameters are dependent across levels, and observations exhibit varying degrees of exchangeability that need to be accounted for (i.e., observations are exchangeable within groups but not across groups). In particular, the following additions are necessary to enable multilevel NPE: 

\begin{itemize}
    \item \textbf{Hierarchical summary networks.} Amortizing over both the number of groups and number of observations requires network training on datasets with varying group sizes and numbers of observations per group. Because observations are nested, hierarchical summary networks are necessary to provide both local information (i.e., characteristics of observations within groups) and global information (i.e., characteristics across groups).
    \item \textbf{Hierarchical inference networks.} In MLMs, the dimension of the joint posterior distribution depends on the number of groups and may also vary between datasets, which requires inference networks with varying number of outputs. This can be realized by hierarchical inference networks.
    \item \textbf{Different types of summary and inference networks.} Multilevel models feature different types of dependencies. For example, hierarchical time series models have non-exchangeable observations within groups, while other model classes may require exchangeability. Neural network architectures for estimating multilevel models therefore need to be flexible enough to account for these differences. 
\end{itemize}

In what follows, we present a neural architecture that is based on NPE with learned summary statistics and enables fully amortized Bayesian inference for MLMs with both tractable and intractable likelihoods. We call this method \textit{Multilevel Neural Posterior Estimation (ML-NPE)}.

\paragraph{Model structure}
We consider a two-level hierarchical model as introduced in \autoref{sec:notation} with local parameters $\lambda_j$, hyperparameters $\tau$ and shared parameters $\omega$. The observation with index $i \in 1,\dots,N_j$ within group $j \in 1,\dots,J$ is denoted by $x_{ij}$, and the data of group $j$ is written as $x_j$. The entire dataset is represented by $x$ with $x=\{x_j\}_{j=1}^{J}$. For visual clarity, we omit the indexing ranges for local parameters and denote the set of all local parameters $\{\lambda_j\}_{j=1}^{J}$ simply as $\{\lambda_j\}$.

Building on the approach of \cite{heinrich_hierarchical_2023}, which focuses on applications in the physical sciences, we extend amortized inference to multilevel models by addressing additional challenges such as varying group sizes and number of observations per group, covariates, and non-IID responses. We leverage the dependency structure implied by the probabilistic graph shown in \autoref{fig:twolevel-graph-and-architecture} (left panel). The joint distribution $p(\tau, \omega, \{\lambda_j\}, x)$ factorizes as:
\begin{equation}
    p(\tau, \omega, \{\lambda_j\}, x) = p(\tau)p(\omega)\prod_{j=1}^{J}p(\lambda_j \given \tau)p(x_{j} \given \lambda_j,\omega)
\end{equation}
This means that the joint posterior 
\begin{equation}
    p(\tau, \omega, \{\lambda_j\} \given x) = \frac{p(\tau, \omega, \{\lambda_j\}, x)}{p(x)} 
\end{equation}
can be further factorized into two parts, namely (1) the hyperparameters $\tau$ together with the shared parameters $\omega$ conditional on all data; and (2) the local parameters $\lambda_j$ conditional on the global parameters and the group-level data:
\begin{align}
    p(\tau, \omega, \{\lambda_j\} \given x) &= p(\tau, \omega \given x) \; p(\{\lambda_j\} \given \tau, \omega, x)\\
    \label{eq:ml-posterior-factorization}
    &= p(\tau, \omega \given x) \prod_{j=1}^{J}p(\lambda_j \given \tau, \omega, x_{j})
\end{align}
This factorization suggests a two-stage approach: First, the global parameters $\tau$ and $\omega$ are estimated conditionally on all observed data. Then, the local parameters $\lambda_j$ are estimated conditionally independent of each other given $\tau$, $\omega$ and the group-level data $x_j$.
In this way, we divide the problem of learning the joint posterior into two smaller problems of learning the global parameters $\tau$ and $\omega$, and the local parameters $\lambda_j$ separately. The factorization of the posterior in \autoref{eq:ml-posterior-factorization}
is not unique and we discuss another possible factorization in \autoref{sec:supp:alternative-factorization}.

\paragraph{Network architecture}
To represent this posterior, we introduce a \textit{hierarchical summary network}, which consists of a series of two summary networks of arbitrary architectures. The first summary network $h_{\psi_{\text{local}}}(x_j)$ transforms the group-level data $x_j$ (with potentially different sample sizes) into fixed-length representations, whereas another summary network transforms the output of the first summary network for all groups into a fixed-length representation of the whole dataset: $h_{\psi_{\text{global}}}(\{h_{\psi_{\text{local}}}(x_j)\}_{j=1}^{J})$.
The architecture of the local summary network $h_{\psi_{\text{local}}}$ can be freely chosen according to the structure of the group-level data (e.g., recurrent neural networks for time series), while the global summary network $h_{\psi_\text{global}}$ is typically chosen to account for exchangeability of groups, for example, via permutation-invariant neural networks \citep[e.g.,][]{zaheer2017deep, lee2019set}.  

Similarly, we also use two inference networks to represent the factorization of the posterior in \autoref{eq:ml-posterior-factorization}. A global inference network $f_{\phi_{\text{global}}}(\tau,\omega; h_{\psi_{\text{global}}}(\{h_{\psi_{\text{local}}}(x_j)\}_{j=1}^{J}))$ estimates the global parameters conditional on the output of the global summary network, while a local inference network $f_{\phi_{\text{local}}}(\lambda_j; \tau, \omega, h_{\phi_{\text{local}}}(x_j))$ estimates the local parameter for group $j$ conditional on the global parameters and the output of the local summary network for inputs $x_j$. \autoref{fig:twolevel-graph-and-architecture} (right panel) shows a diagram of the network architecture.

\paragraph{Network training}
Writing $\phi=(\phi_{\text{global}}, \phi_{\text{local}})$ for the weights of the inference networks and $\psi=(\psi_{\text{global}}, \psi_{\text{local}})$ for the weights of the summary networks, the optimization criterion of \autoref{eq:npe-loss-joint} becomes
\begin{equation}
    \widehat{\phi}, \widehat{\psi} = \argmin_{\phi, \psi} \mathbb{E}_{(\tau, \omega, \theta, x) \sim p(\tau, \omega, \theta, x})\Big[-\log q_{\phi, \psi}\big(\tau, \omega,\{ \lambda_j\}\given x \big)\Big]
\end{equation}
The negative log-density term inside the expectation can now be decomposed into:
\begin{equation}
    \begin{split}
        -\log q_{\phi}\big(\tau, \omega,\{\lambda_j\}\given x \big) = &-\log q_{\phi_{\text{global}}}(\tau, \omega \given h_{\psi_{\text{global}}}(\{h_{\psi_{\text{local}}}(x_j)\}_{j=1}^{J})) \\
        & -\sum_{j=1}^{J}\log q_{\phi_{\text{local}}}(\lambda_j \given \tau, \omega, h_{\psi_{\text{local}}}(x_j)),
    \end{split}
\end{equation}
which can be optimized via standard backpropagation.

\paragraph{Amortized sampling from ML-NPE}
After training, the neural networks can be saved and reused for posterior inference on new datasets with any number of groups $J$ and number of observations per group $N_j$, under the condition that they remain within the typical range seen during training.

Let $z_{f_{\phi_{\text{global}}}}^{(s)}$ and $z_{f_{\phi_{\text{local}}}}^{(s)}$ denote random draws with the index $s \in \{1,\dots,S\}$ from the base distribution of the global inference network $f_{\phi_{\text{global}}}$ and local inference network $f_{\phi_{\text{local}}}$, respectively (here: both unit Gaussian, see \autoref{sec:npe} for details). Let $x^{\text{obs}}$ denote an observed dataset. Then, random draws from the approximate posterior $q_{\phi,\psi}(\tau, \omega, \{\lambda_j\} \given x^{\text{obs}})$ can be obtained by the following ancestral sampling scheme:
\begin{equation}
    \begin{split}
        z_{\phi_{\text{global}}}^{(s)} &\sim \text{Normal}(z_{\phi_{\text{global}}} \given 0, \mathbb{I}) \\
        z_{\phi_{\text{local}}}^{(s)} &\sim \text{Normal}(z_{\phi_{\text{local}}} \given 0, \mathbb{I}) \\
       \tau^{(s)}, \omega^{(s)} &= f^{-1}_{\phi_{\text{global}}}(z_{\phi_{\text{global}}}^{(s)};h_{\psi_{\text{global}}}(\{h_{\psi_{\text{local}}}(x^{\text{obs}}_j)\}_{j=1}^{J})) \\
        \lambda_j^{(s)} &= f^{-1}_{\phi_{\text{local}}}(z_{\phi_{\text{local}}}^{(s)};\tau^{(s)}, \omega^{(s)},h_{\psi_{\text{local}}}(x^{\text{obs}}_j)) \quad \textbf{for}\;j=1,\dots,J,
   \end{split}
\end{equation}
where $f^{-1}_{\phi_{\text{global}}}$ and $f^{-1}_{\phi_{\text{local}}}$ denote inverse passes through the respective invertible inference networks. This approach ensures efficient sampling at each level, requiring only single passes through the pre-trained invertible networks. Additionally, using only a single inference network for the local parameters $\{\lambda_j\}$ prevents the output dimensionality of the local inference network from scaling linearly with the number of groups $J$. 


\section{Empirical evaluation} \label{sec:experiments}

We evaluate our neural two-level architecture in three different applications based on real-world datasets that aim to cover a broad spectrum of use cases for multilevel models. All models are implemented in the BayesFlow software for amortized Bayesian workflows \citep{radev2023joss}. Concretely, the applications include the following: 

\begin{enumerate}
    \item An autoregressive time series model that predicts and compares air passenger traffic between European countries and the United States of America. 
    \item A diffusion decision model to infer latent parameters of a decision-making process from reaction time data. 
    \item A generative neural network model to infer handwriting styles from high-dimensional image data. 
\end{enumerate}

\subsection{Evaluation metrics}
For each application, we implement a series of comprehensive model checks: 

\paragraph{Simulation-based calibration.}
We perform simulation-based calibration \citep[SBC, e.g.,][]{Cook2006, Talts2018, modrak2023simulation} to ensure that ML-NPE samples from the correct posterior. SBC assesses correct posterior calibration (i.e., a $q\times100\%$ posterior interval contains the true value in approximately $q\times100\%$ of the cases) \citep[e.g.,][]{buerkner2023taxonomy}. Incorrect calibration hints at issues with computational validity, for example arising from insufficient simulation-based training or lack of expressiveness of the neural networks.

\paragraph{Posterior predictive checks.}
To ensure that all models are reasonable descriptions of the observed data, we perform posterior predictive checks \citep[e.g.,][]{gabry2019visualization}. We investigate inferential accuracy by providing graphical comparisons of posterior intervals and ground truth parameter values over a test set. 
When computationally feasible, we also fit each model in Stan and provide plots of posterior intervals to compare the estimates against a known to be reliable reference algorithm. 
These model tests on non-simulated data are particularly important in amortized Bayesian inference because of a possible amortization gap: If the simulated data used for training does not lie within the scope of data the model is going to be applied on, ABI might yield posterior draws that are far off from the true posterior.

\paragraph{Posterior shrinkage.}
As a final model check, we compare posterior shrinkage of the group-level parameters toward their global mean between our neural framework and Stan, as any subtle algorithmic error would likely first manifest as incorrect posterior shrinkage. Shrinkage is achieved by placing a hierarchical prior on the local parameters, such as $\eta_j \sim \text{Normal}(\mu_\eta, \sigma_\eta)$, where $\eta_j$ is an individual local parameter and $\mu_\eta$ and $\sigma_\eta$ are the hierarchical mean and standard deviation for that parameter, respectively. This prior structure pulls individual estimates $\eta_j$ closer together, reducing variance and borrowing strength across groups \citep[e.g.,][]{efron1975data, carlin1997bayes, Gelman2006}. Following \citet{Gelman2006a}, a useful way to quantify this effect is the pooling factor $\kappa_j$:
\begin{equation}
    \kappa_j=\frac{\text{Var}(\eta_j - \mu_\eta)^2}{\sigma_{\eta}^2},
\end{equation}
where $\text{Var}(\cdot)$ denotes the variance and $\sigma_{\eta}^2$ is the squared hierarchical standard deviation. 

\subsection{Experiment 1: Air passenger traffic} \label{subsec:air-traffic}

We apply ML-NPE to analyze trends in European air passenger traffic data provided by \citet{Eurostat2022b, Eurostat2022a, Eurostat2022c}. We highlight that our neural architecture can correctly and efficiently amortize over both the number of groups and the number of observations per group, as well as over covariates. By using a summary network that is aligned to the structure of the data, we also show that ML-NPE can easily be extended to estimate non-IID response variables.

\paragraph{Model description}
We obtain time series of annual air passenger counts between 16 European countries (departures) and the United States of America (destination) from 2004 to 2019 and fit the following autoregressive multilevel model:
\begin{equation}
    y_{t+1,j} \sim \mathrm{Normal}(\alpha_j + y_{t,j}\beta_j + u_{t,j}\gamma_j + v_{t,j}\delta_j, \sigma_j),
    \label{eq:air-traffic-autoregressive}
\end{equation}
where the target quantity $y_{t+1,j}$ is the difference in air passenger traffic for country $j$ between time $t+1$ and $t$. To predict $y_{j,t+1}$, we use two additional predictors: The first predictor $u_{t,j}$ is the annual household debt of country $j$ at time $t$, measured in \% of gross domestic product (GDP), and the second predictor $v_{t,j}$ is the real GDP per capita.
The parameters $\alpha_j$ are the country-level intercepts, $\beta_j$ are the autoregressive coefficients, $\gamma_j$ are the regression coefficients of the household debt predictor, $\delta_j$ are the regression coefficients for the GDP per capita predictor, and $\sigma_j$ is the standard deviation of the noise term. 

Mapping the notation to the previous sections, the observables $x_j$ are now split into a target $y_j=\{y_{t,j}\}_{t=1}^{T}$ and predictors $u_{j}=\{u_{j,t}\}_{t=1}^{T}$ and $v_{j}=\{v_{j,t}\}_{t=1}^{T}$, with groups defined by the country index $j$.
We assume independent priors and use the following marginal prior distributions:
\begin{flalign}
\begin{aligned}
\begin{split}
    \alpha_j &\sim \text{Normal}(\mu_\alpha, \sigma_\alpha) \\
    \beta_j &\sim \text{Normal}(\mu_\beta, \sigma_\beta) \\
    \gamma_j &\sim \text{Normal}(\mu_\gamma, \sigma_\gamma) \\
    \delta_j &\sim \text{Normal}(\mu_\delta, \sigma_\delta) \\
    \log(\sigma_j) &\sim \text{Normal}(\mu_\sigma, \sigma_\sigma) \\
\end{split}
\hspace{0.5cm}
\begin{split}
    \mu_\alpha &\sim \text{Normal}(0, 0.5) \\
    \mu_\beta &\sim \text{Normal}(0, 0.2) \\
    \mu_\gamma &\sim \text{Normal}(0, 0.5) \\
    \mu_\delta &\sim \text{Normal}(0, 0.5) \\
    \mu_\sigma &\sim \text{Normal}(-1, 0.5) \\
\end{split}
\hspace{0.5cm}
\begin{split}
    \sigma_\alpha &\sim \text{Normal}^{+}(0, 0.25) \\
    \sigma_\beta &\sim \text{Normal}^{+}(0, 0.15) \\
    \sigma_\gamma &\sim \text{Normal}^{+}(0, 0.25) \\
    \sigma_\delta &\sim \text{Normal}^{+}(0, 0.25) \\
    \sigma_\sigma &\sim \text{Normal}^{+}(0, 1) \\
\end{split}
\end{aligned}
\label{eq:air-traffic-generative-model-hypers}
\end{flalign}
As commonly done for autoregressive models, we regress on the differences between time periods to reduce issues due to non-stationarity. This is particularly important in an SBI setting, because for $\beta_j > 1$, the time series would exhibit strong exponential growth that would quickly surpass reasonable air traffic volumes, creating highly unrealistic simulations.

\paragraph{Model training}
To generate training data for the neural approximator, we utilize an ancestral sampling scheme: We define the set of hierarchical means as $\mu=\{\mu_\alpha, \mu_\beta, \mu_\gamma, \mu_\delta, \mu_\sigma\}$ and the set of hierarchical standard deviations as $\sigma=\{\sigma_\alpha, \sigma_\beta, \sigma_\gamma, \sigma_\sigma\}$, which together form the set of hyperparameters $\tau=\{\mu, \sigma\}$. The local parameters are given by $\lambda_j=\{\alpha_j, \beta_j, \gamma_j, \delta_j, \sigma_j\}$ and the model does not have any shared parameters.

We generate training data indexed by $m$, where $m \in \{1,\dots,M\}$. For each training sample $m$, we first draw a set of hyperparameters from the prior distribution:
\begin{equation}
    \tau^{(m)} \sim p(\tau).
\end{equation}
Next, we sample $J_m$ group-level parameters from the conditional distribution:
\begin{equation}
    \{\lambda_j^{(m)}\}_{j=1}^{J_m} \given \tau^{(m)} \sim p(\lambda_j \given \tau^{(m)}),
\end{equation}
where $J_m$ is the total number of groups in the $m$th training sample, and each index j corresponds to a country. For each country, we then simulate a time series according to \autoref{eq:air-traffic-autoregressive}. To facilitate training on simulated data and without loss of generality, we assume that each time series begins with an initial value drawn from
\begin{equation}
    y_{0,j}^{(m)} \sim \text{Normal}(0.5, 1).
\end{equation}
Additionally, the predictor variables $u_j^{(m)}$ and $v_j^{(m)}$ are sampled from a standard normal distribution and then standardized to have a mean of 0 and standard deviation 1.
Defining
\begin{equation}
    x_j^{(m)}=\{y_j^{(m)}, u_j^{(m)}, v_j^{(m)}\}
\end{equation}
and suppressing the indexing ranges from $j=1$ to $J_m$ for visual clarity, the final training dataset consists of
\begin{equation}
    \{\tau^{(m)}, \{\lambda_j^{(m)}\},\{x_j^{(m)}\}\}_{m=1}^{M}.
\end{equation}
To amortize over the number of countries and observed time points, we vary $J_m \sim \text{DiscreteUniform}(10, 30)$ and $T_m \sim \text{DiscreteUniform}(5, 30)$ for each training sample $m$. The neural approximator can therefore generate draws from the posterior $p(\tau, \{\lambda_j\} \given \{x_j\})$ for datasets with a varying number of countries and a varying number of time points without further training.

Network configuration details are provided in \autoref{sec:supp:training-details}. Briefly, the local summary network has an initial long short-term memory \citep[LSTM;][]{Hochreiter1997} layer to capture time dependencies at the group level. For the global summary network, we employ a set transformer \citep{Lee2018} to ensure exchangeability of groups. The inference networks are implemented as neural spline flows \citep{Durkan2019}. We set the simulation budget to $M=\num{100000}$ and train for 200 epochs using the Adam optimizer \citep{Kingma2014}.

\begin{figure}
    \centering
    \includegraphics[trim={0, 0, 0, 15mm}, clip, width=\linewidth]{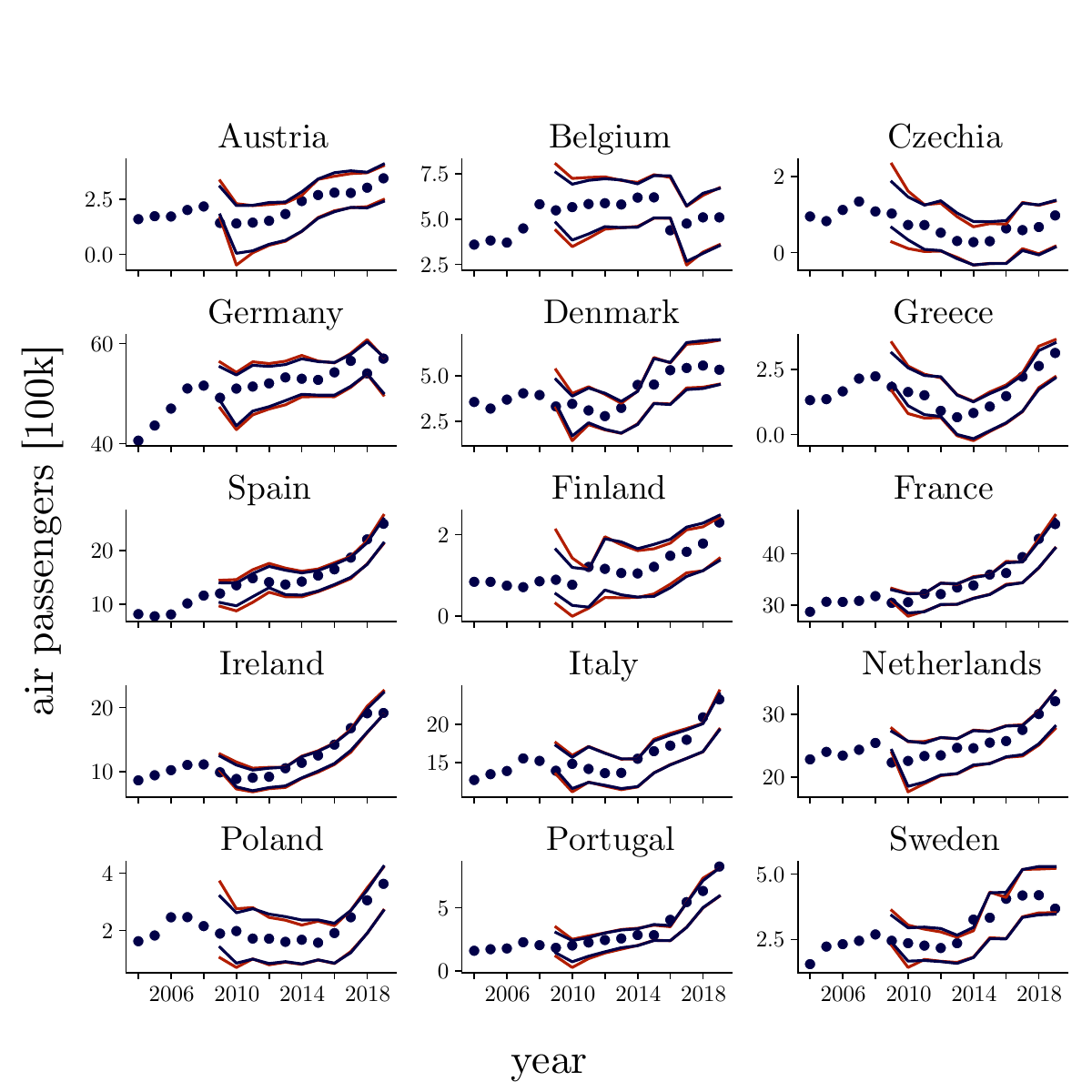}
    \caption{Annual air passenger traffic volume in 100 thousand passengers between different European countries and the United States of America. Blue points show observed data, blue lines show 95\% uncertainty intervals of the 1-step ahead posterior predictive distribution. For reference, red lines show draws from the posterior predictive distribution of the same model fitted in Stan. The posterior predictive checks show that the model predictions are consistent with the observed data and intervals of the posterior predictive distribution obtained by the reference implementation in Stan are mostly indistinguishable from prediction intervals obtained by our amortized ML-NPE method. Estimated differences in year-to-year air passenger counts are displayed as 1-step ahead predictions.}
    \label{fig:air-traffic-ppc}
\end{figure}

\begin{figure}
    \centering
    \includegraphics[width=\linewidth]{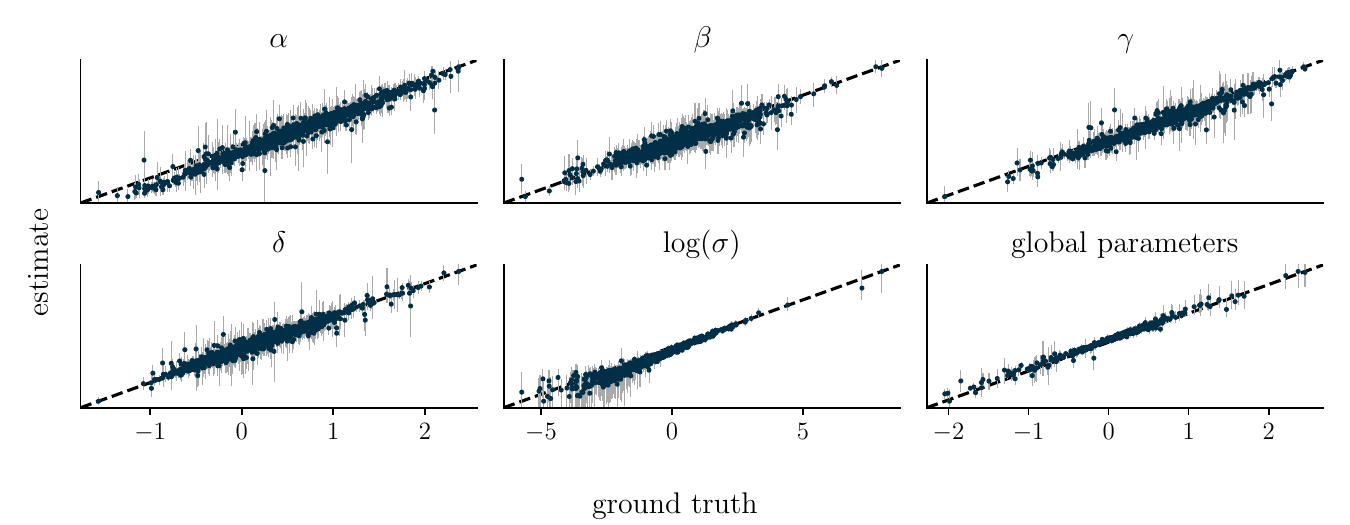}
    \caption{Marginal posterior estimates vs. ground truth on 50 simulated validation datasets. All parameters show good recovery of the true values, indicating that our ML-NPE is able to recover true parameter values without bias.}
    \label{fig:air-traffic-recovery}
\end{figure}

\begin{figure}
    \centering
    \includegraphics[width=\linewidth]{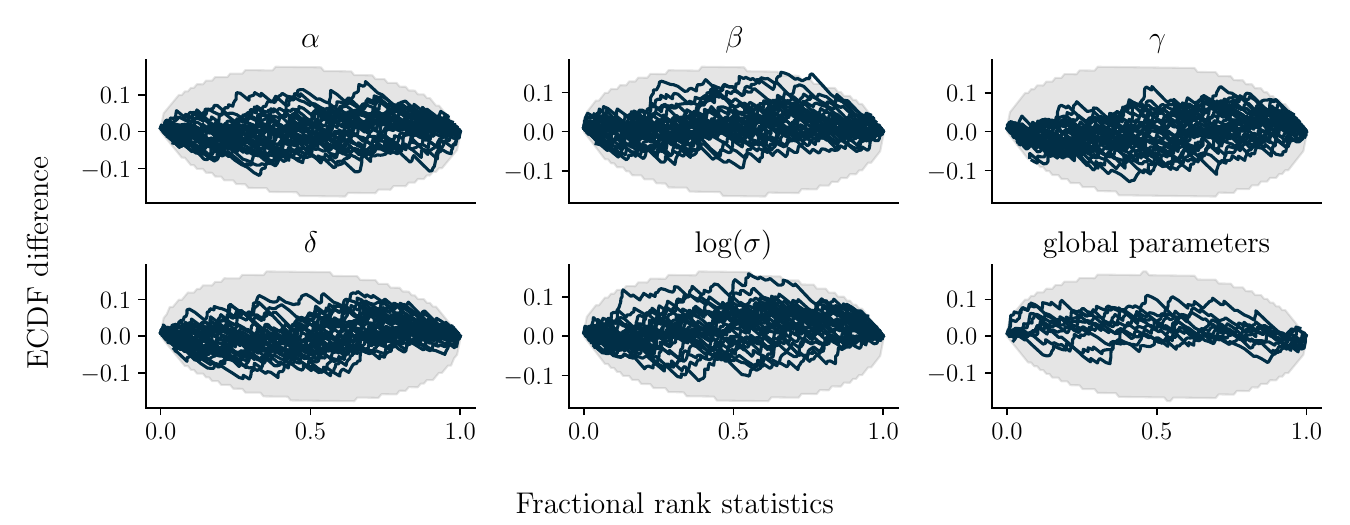}
    \caption{Simulation-based calibration plots based on 100 simulated validation sets. ECDF difference is the difference between the empirical cumulative distribution function of the rank distribution (obtained by comparing the rank of prior draws with their corresponding posterior draws) and the uniform cumulative distribution function \citep[see][for details]{sailynoja2022graphical}. For the parameters $\alpha$, $\beta$, $\gamma$, $\delta$, and $\text{log}(\sigma)$, each line represents a group-level parameter (i.e., the country-specific estimate). The panel \emph{global parameters} contains the estimated mean and standard deviation for each of the 5 group-level parameters. All lines lie within the shaded 99\% simultaneous confidence bands \citep{sailynoja2022graphical}, indicating well-calibrated marginal posterior distributions.}
    \label{fig:air-traffic-sbc}
\end{figure}

\begin{figure}
\centering
\includegraphics[width=0.95\linewidth]{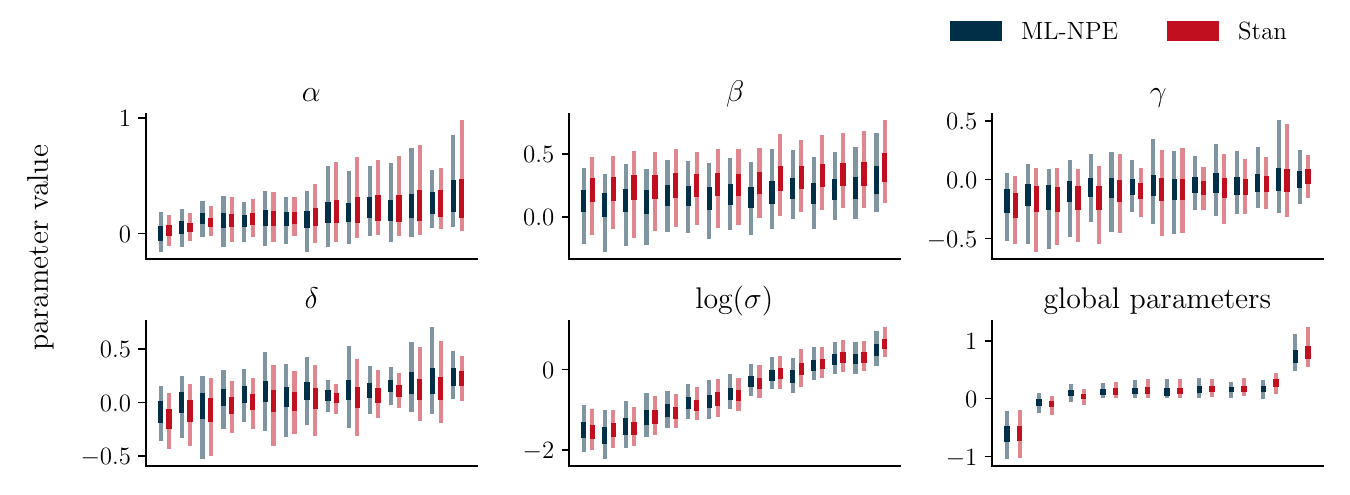}
\caption{Posterior intervals obtained from amortized ML-NPE (ours) and the gold-standard HMC reference, as implemented via Stan (computed on non-simulated data). 
Plots show the central 50\% (dark) and 95\% (light) posterior credible intervals based on quantiles.
For the parameters $\alpha$, $\beta$, $\gamma$, $\delta$, and $\text{log}(\sigma)$, each interval pair refers to a single group-level parameter (one for each country). For the global parameters, each interval pair refers to a hyperparameter (one mean and one standard deviation for each of the group-level parameters, so 10 hyperparameters in total). The parameters are sorted by increasing mean (as per Stan) to ease interpretation of shrinkage towards a common mean.}
\label{fig:air-traffic-posterior}
\end{figure}

\paragraph{Posterior predictive checks}
To confirm that  the simple AR(1) model is adequate enough to describe the observed data, \autoref{fig:air-traffic-ppc} shows the observed time series overlaid with draws from the posterior predictive distribution obtained by our ML-NPE method as well as a reference implementation in Stan. While predictive uncertainty is high, presumably in part because of the relatively simple model, short time series and predictors that are only weakly associated with the response variable, the predictions from the AR(1) model are consistent with the observed data.
\paragraph{Posterior inference}
\autoref{fig:air-traffic-recovery} compares marginal posterior estimates with simulated ground truths (posterior recovery) and \autoref{fig:air-traffic-sbc} shows posterior calibration on 100 simulated validation datasets. We observe good posterior recovery and calibration for all parameters.
Further, to assess posterior recovery on real data, we also apply our model to the \citet{Eurostat2022b, Eurostat2022a, Eurostat2022c} dataset and compare the results to the posterior intervals obtained by the Stan reference model implementation (see \autoref{fig:air-traffic-posterior}). The plot shows that the results obtained by ML-NPE are consistent with the results obtained by Stan. 


\begin{figure}
    \centering
    \includegraphics[width=\linewidth]{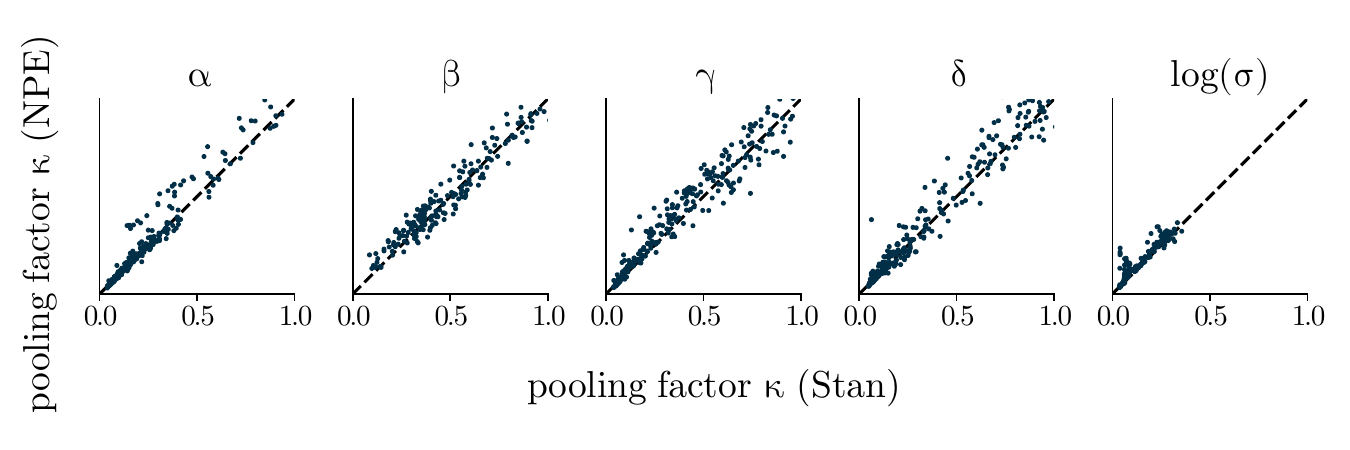}
    \caption{Posterior shrinkage observed in 100 simulated datasets via our ML-NPE methods to the results obtained by Stan. The scatter points lie along the diagonal, indicating that our method is able to correctly shrink group-level regression coefficients towards their common mean.}
    \label{fig:air-traffic-shrinkage}
\end{figure}

\paragraph{Posterior shrinkage}
To verify correct shrinkage behavior of the ML-NPE approach, we compare posterior shrinkage to a reference implementation in Stan.
\autoref{fig:air-traffic-shrinkage} shows scatter plots of shrinkage factors for each parameter obtained on 100 simulated datasets via our ML-NPE method compared to shrinkage factors obtained on the same data via Stan. The scatter points lie along the diagonal, indicating correct shrinkage behavior of our ML-NPE method.

\subsection{Experiment 2: Diffusion Decision Model}

As a second experiment, we apply ML-NPE to a popular decision-making model in psychology, neuroscience, and the cognitive sciences. We use this experiment to highlight that NPE is not only well-suited for standard use cases like real-time analysis or fitting the same model across multiple datasets, but also for following a traditional Bayesian workflow, e.g. when performing model checking using cross-validation. Additionally, we show that ML-NPE can reliably amortize over both the number of groups and observations, enabling fast and reliable posterior inference for almost arbitrary datasets in this model class.

\begin{figure}
    \centering
    \includegraphics[width=0.77\linewidth]{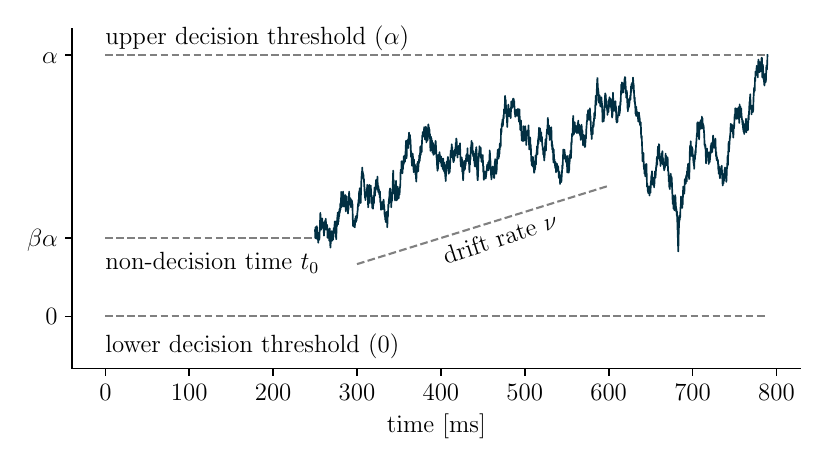}
    \caption{Example of a latent evidence trajectory for a single subject. The decision making process is modeled as a random walk that is governed by 4 parameters: The non-decision time $t_0$ is the time until evidence accumulation begins and captures components that are not directly related to the decision, such as time spent on processing of sensory information. The drift rate $\nu$ corresponds to the rate of information uptake. The decision threshold $\alpha$ can be interpreted as the amount of evidence necessary to make a decision and is thus a measurement of response caution. The starting point $\beta$ quantifies a subject's tendency to prefer one choice over the other.}
    \label{fig:ddm-info}
\end{figure}

\paragraph{Model description} \label{sec:ddm-model-description}
Consider a decision task in which participants are presented with sequences of letters and asked to differentiate between words and non-words (i.e., a lexical decision task). The Diffusion Decision Model \citep[DDM; e.g.,][]{ratcliff2016diffusion} simulataneously models this binary decision and the response time via a continuous evidence accumulation process: After an initial non-decision time $t_0$, evidence accumulates following a noisy diffusion process with a certain drift rate $\nu$, starting from a point $\beta$, until one of two decision thresholds $\{0, \alpha\}$ corresponding to the two choices is hit (see \autoref{fig:ddm-info}).
The nested structure of individual observations within subjects lends itself to a multilevel model where subject-specific estimates share information via common hyper-priors. For the first part of this experiment, we consider subject-specific estimates for the non-decision time $t_{0,j}$, drift rate $\nu_j$ and decision threshold $\alpha_j$ (local parameters), while estimating a global starting point $\beta$ for all subjects (shared parameter).
The generative model is summarized as:
\begin{flalign}
\begin{aligned}
\nu_j &\sim \text{Normal}(\mu_\nu, \sigma_\nu) \\
\alpha_j &\sim \text{Normal}(\mu_\alpha, \sigma_\alpha) \\
t_{0,j} &\sim \text{Normal}(\mu_{t_0}, \sigma_{t_0}) \\[0.5cm]
\beta &\sim \text{Beta}(50, 50) \\
y_j &\sim \text{DDM}(\nu_j, \alpha_j, t_{0,j}, \beta),
\end{aligned}
\hspace{0.5cm}
\begin{aligned}
\mu_\nu &\sim \text{Normal}(0.5, 0.3) \\
\mu_\alpha &\sim \text{Normal}(0, 0.05) \\
\mu_{t_0} &\sim \text{Normal}(-1, 0.3) \\[0.5cm]
& \\
&
\end{aligned}
\hspace{0.5cm}
\begin{aligned}
\log\sigma_\nu &\sim \text{Normal}(-1, 1) \\
\log\sigma_\alpha &\sim \text{Normal}(-3, 1) \\
\log\sigma_{t_0} &\sim \text{Normal}(-1, 0.3) \\[0.5cm]
& \\
&
\end{aligned}
\end{flalign}
where $y_j$ is the vector containing tuples of binary decisions and corresponding reaction times for subject~$j$, $\text{DDM}$ is the likelihood of the 4-parameter diffusion decision model, $\beta$ is the shared bias for all subjects, $\text{Beta}$ is the Beta distribution with shape parameters $a$ and $b$, $\text{Normal}$ is the normal distribution with mean $\mu$ and standard deviation $\sigma$, and $\nu_j$, $\alpha_j$, $t_{0,j}$ are the subject-specific drift rate, boundary threshold and non-decision time parameters, respectively.
The priors were chosen based on prior predictive checks to match the range of typical outcomes for such decision-making tasks.

\paragraph{Model training}
To generate data for network training, we use the same ancestral sampling scheme as for the air traffic experiment in \autoref{subsec:air-traffic}. First, we generate $m \in \{1\dots,M\}$ random draws from the hyperparameters $\tau=\{\mu_\nu,\mu_\alpha,\mu_{t_0},\sigma_\nu,\sigma_\alpha,\sigma_{t_0}\}$ and shared parameters $\omega=\{\beta\}$. For each of these $M$ random draws, we then draw $J_m$ draws from the corresponding subject specific local parameters $\lambda_j=\{\nu_j,\alpha_j,t_{0,j}\}$. For each subject, we  simulate $N_m$ observations from the diffusion process. To facilitate amortization over different group and observation sizes, we randomly draw values of $J_m$ from a discrete uniform distribution with lower and upper bounds of $10$ and $30$, and $N_m$ from a discrete uniform distribution with lower and upper bounds of $1$ and $100$. 
Details of the network configuration are shown in \autoref{sec:supp:training-details}. Briefly, we use set transformers \citep{Lee2018} for both the local and global summary networks and neural spline flows \citep{Durkan2019} for both inference networks. The simulation budget is set to $M=\num{20000}$ and we train for 200 epochs using the Adam optimizer \citep{Kingma2014}. 

\begin{figure}
     \centering
     \includegraphics[width=0.95\linewidth]{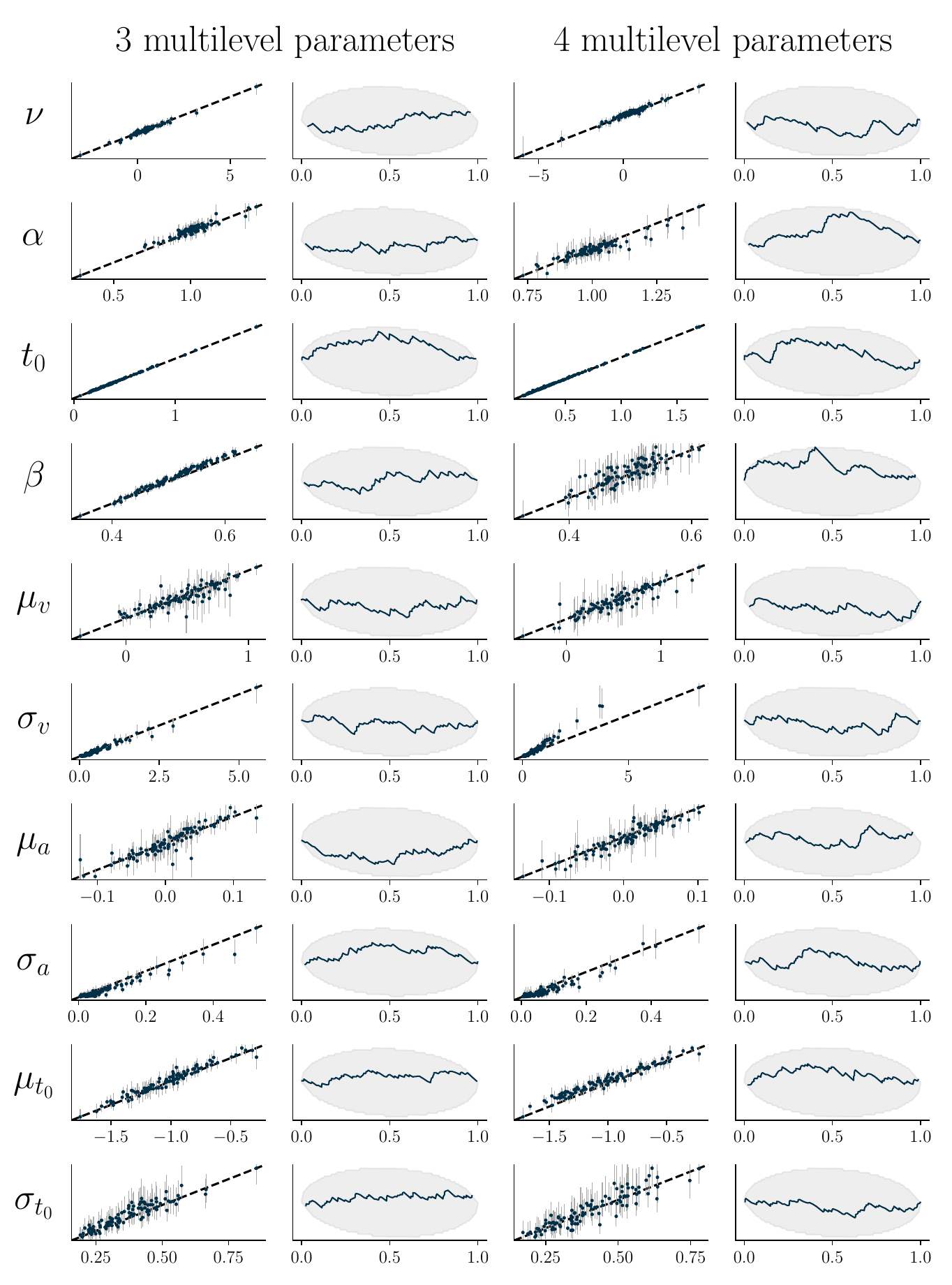}
     \caption{Posterior recovery (columns 1 and 3) and calibration (columns 2 and 4) for the diffusion decision model on 100 simulated validation datasets. The first two columns correspond to a model where a single $\beta$ is estimated for all subjects (shared parameter). The last two columns correspond to a model where $\beta$ is also estimated hierarchically. All parameters show good posterior recovery and calibration}
     \label{fig:posterior-recovery-calibration}
 \end{figure}

 \paragraph{Posterior inference} \label{sec:ddm-posterior-inference}
\autoref{fig:posterior-recovery-calibration} shows posterior recovery and calibration on 100 simulated validation datasets. We observe good posterior recovery and calibration for all parameters.

\begin{figure}
     \centering
    \includegraphics[width=\linewidth]{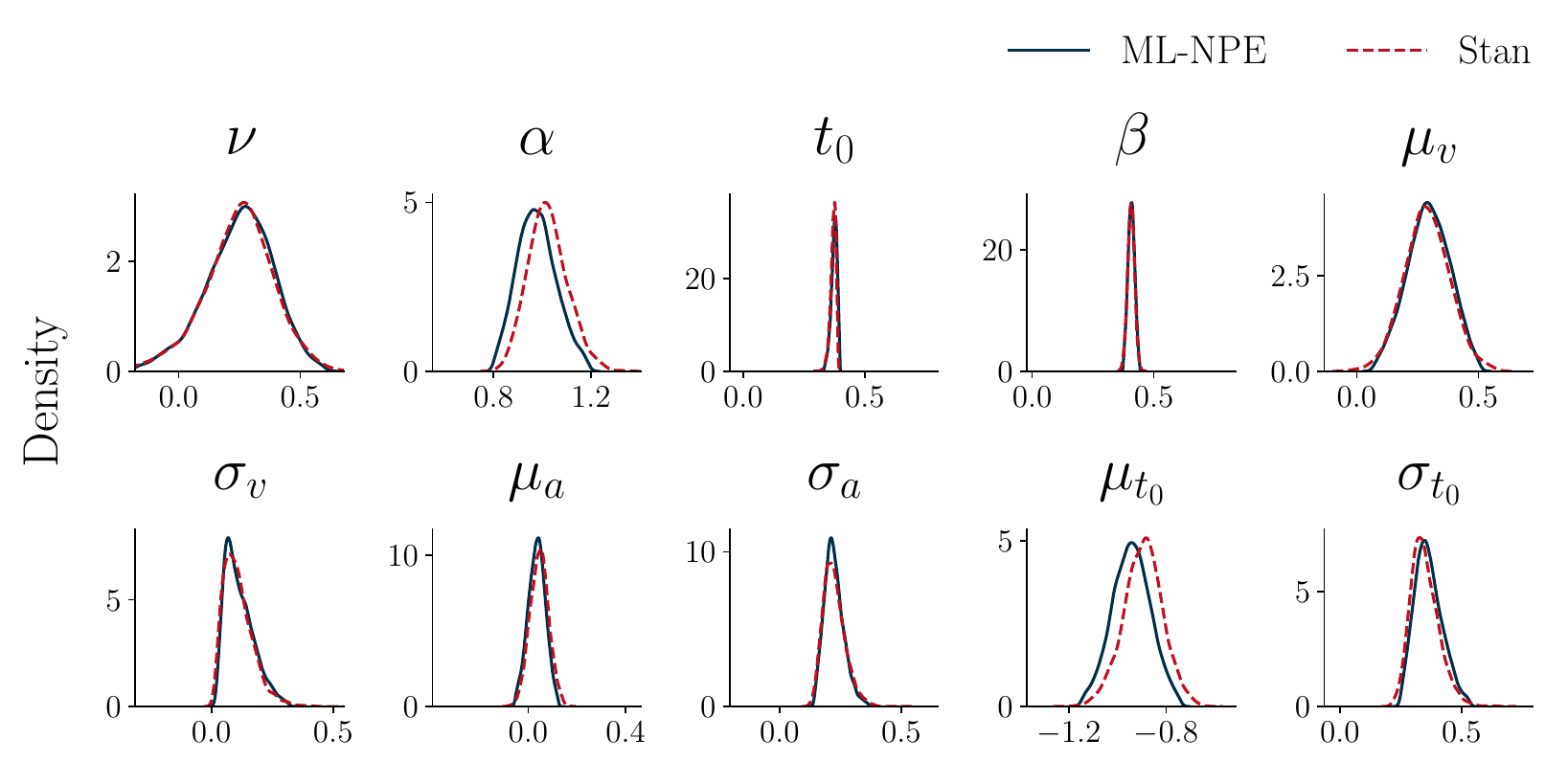} 
     \caption{Marginal posterior distributions of the drift diffusion model fitted on experimental data. Blue kernel density plots were obtained by our ML-NPE method, red dashed lines show the results obtained by Stan as a gold standard for reliable posterior inference. The plots show that marginal posteriors obtained by ML-NPE are highly similar to the marginal posteriors obtained by Stan.}
     \label{fig:ddm-marginal-posteriors-vs-stan}
 \end{figure}

To validate that we also observe accurate posterior inference on non-simulated data, we contrast these results to model fits on experimental data published by \citet{Wagenmakers2008}. In total, 17 subjects were asked to make 32 word/non-word decisions in each of the 20 trial blocks. The experiment was repeated in two conditions: once after asking the subjects to focus on accuracy and once after asking the subjects to focus on speed. 
\autoref{fig:ddm-marginal-posteriors-vs-stan} shows marginal posterior densities for the model fit on a subset of the data (observations in the accuracy condition in trial block 4) compared to the results obtained by Stan.

\paragraph{Leave-one-group-out cross-validation} \label{sec:ddm-logo}

Amortized Bayesian inference is not only useful when performing posterior inference on many different datasets. Refitting the model multiple times on different subsets of the data is also required for many essential steps in the modern Bayesian workflow \citep{Gelman2020}, such as cross-validation \citep[e.g,][]{Vehtari2017, merkle_bayesian_2019}.
For multilevel models, it is often of practical interest to estimate and compare the predictive performance of models on new groups, leading to leave-one-group-out (LOGO) cross-validation.
We define the leave-one-group-out posterior as 
\begin{equation}
p(\theta \given \{x_{-j}\}) = \frac{p(\{x_{-j}\} \given \theta)p(\theta)}{p(\{x_{-j}\})}, 
\end{equation}
where $\{x_{-j}\}=x \setminus x_j$ denotes the data with the $j$th group removed.

As the LOGO posterior is often vastly different from the full posterior because the local parameters for the left-out group are only identified via their hyperparameters, approximation methods like Pareto-smoothed importance sampling \citep{Vehtari2024} almost always fail.
Performing LOGO cross-validation then requires refitting the model once for each group in the dataset, quickly rendering model comparison computationally infeasible when relying on, for example, MCMC algorithms for posterior inference. With amortized posterior inference, obtaining the LOGO posterior is almost instant (a fraction of a second for our architectures), as it does not require refitting the model for each data subset.

To show that ML-NPE enables model comparison for multilevel models that are otherwise computationally infeasible, we contrast the previously described diffusion decision model in \autoref{sec:ddm-model-description} with another model variant that also estimates $\beta$ hierarchically. Instead of a single shared $\beta$ for all subjects, this model variant estimates subject-specific $\beta_j$ that share common hyper-priors.
We make the following adjustments compared to the previous description of the diffusion decision model:
\begin{flalign}
\begin{aligned}
\begin{split}
    \beta_j &\sim \text{BetaProportion}(\mu_\beta, \xi_\beta)
\end{split}
\hfill
\begin{split}
    \mu_\beta &\sim \text{Beta}(50, 50)
\end{split}
\hspace{0.5cm}
\begin{split}
    \xi &\sim \text{Gamma}(5, 3),
\end{split}
\end{aligned}
\label{eq:ddm-generative-model-beta}
\end{flalign}
where $\beta_j$ is the bias for subject $j$, $\text{BetaProportion}$ is the Beta distribution parameterized by a mean $\mu_\beta$ and concentration parameter $\xi_\beta$.
The Beta proportion distribution is linked to the (standard) shape parameterization of the Beta distribution via the relation $\alpha=\mu\xi$ and $\beta=(1-\mu)\xi$.
This adapted version of the diffusion decision model also illustrates that ML-NPE can estimate non-normal hyper-prior distributions.

 \begin{figure}
     \centering
     \includegraphics[width=0.8\linewidth]{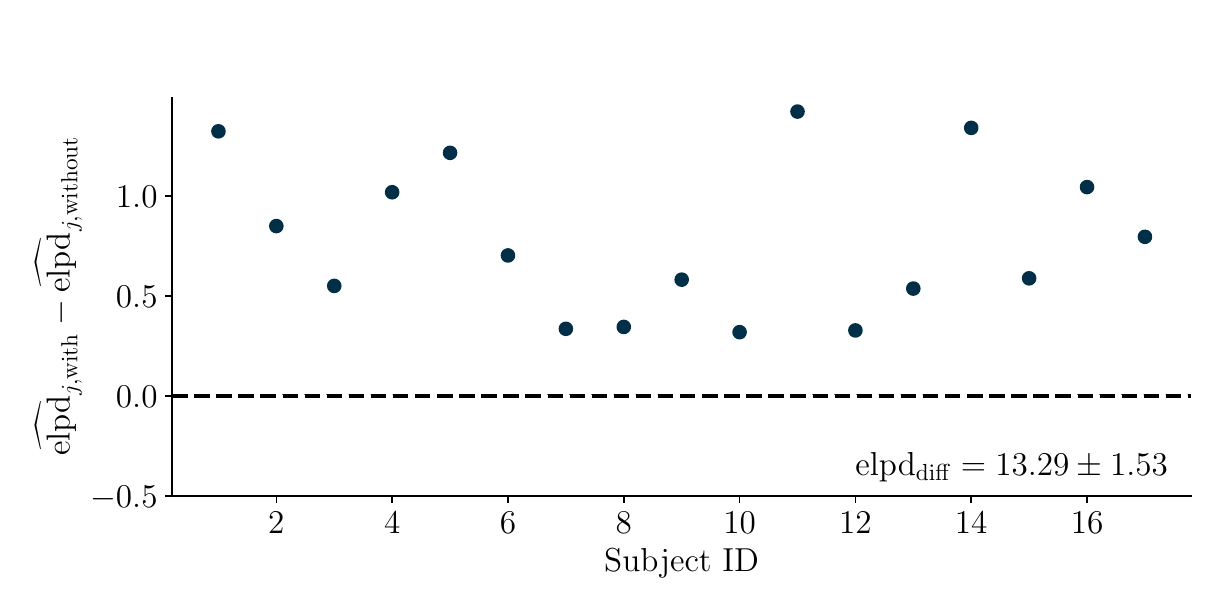}
     \caption{LOGO cross-validation on trial block 4 of the experimental dataset published by \citet{Wagenmakers2008}. 
     Scatter points show subject-wise elpd differences between the models with and without subject-specific $\beta_j$ as defined in \autoref{eq:elpd_j}. The elpd difference $\text{elpd}_{\text{diff}}=\sum_{j=1}^{J}(\widehat{\text{elpd}}_{j,\text{with}} - \widehat{\text{elpd}}_{j,\text{without}})$ indicates a substantially better fit of the more flexible model with 4 local parameters per subject.
     }
     \label{fig:ddm-logo}
 \end{figure}

\autoref{fig:ddm-logo} shows expected log predictive density \cite[elpd;][]{Vehtari2017} values for each subject obtained by first removing subject $j$ from the dataset, generating samples from the LOGO posterior and then evaluating the predictive performance on the left-out subject:
\begin{equation}
    \label{eq:elpd_j}
    \widehat{\text{elpd}_j}=\log(\frac{1}{S}\sum_{s=1}^{S}p(x_j\given \theta^{(s)})),
\end{equation}
where $\theta^{(s)}$ denotes $s=1,\dots,S$ draws from the leave-one-group-out posterior $p(\theta \given \{x_{-j}\})$.

The elpd difference $\text{elpd}_{\text{diff}}=\sum_{j=1}^{J}(\widehat{\text{elpd}}_{j,\text{with}} - \widehat{\text{elpd}}_{j,\text{without}})$ between the models with and without subject-specific $\beta_j$ evaluates to $\text{elpd}_\text{diff} \approx 13.29$, which is several times larger than its standard error $\text{SE}_\text{diff} \approx 1.53$ \citep[][]{Vehtari2017}. That is, there is strong statistical evidence that the model with varying $\beta_j$ has better predictive performance on data of a previously unseen group.

Performing such a model comparison is computationally efficient in an amortized Bayesian inference setting, but quickly becomes infeasible when the model is fit using MCMC-based algorithms. 
In this example, using Stan to draw 1000 samples (after 1000 warm-up draws) from 4 chains in parallel takes about 6 minutes on a standard desktop processor, resulting in a total runtime of about 1.5 hours for all subjects. In contrast, training the amortized model using ML-NPE takes about 1 hour on a standard desktop graphics card, subsequently allowing almost instant inference (a fraction of a second) of the LOGO posteriors. Typical applications of the DDM often feature hundreds of subjects and observations. In these settings, traditional HMC methods can quickly become computationally infeasible even on large compute clusters, but can still be performed on a standard desktop computer using NPE.

This shows that the required training time of amortized methods can quickly pay off, even in scenarios where one is only interested in evaluating a single dataset. This is particularly true if the number of groups is large, with growing gains for amortized methods as the number of groups becomes larger. \autoref{sec:supp:NPE-over-MCMC} includes a brief discussion of use cases where users can benefit from NPE rather than MCMC-based workflows.

\subsection{Experiment 3: Style inference of hand-drawn digits}
\begin{figure}
    \centering
    \includegraphics[width=\linewidth, trim={0 3.6cm 2.5cm 0},clip]{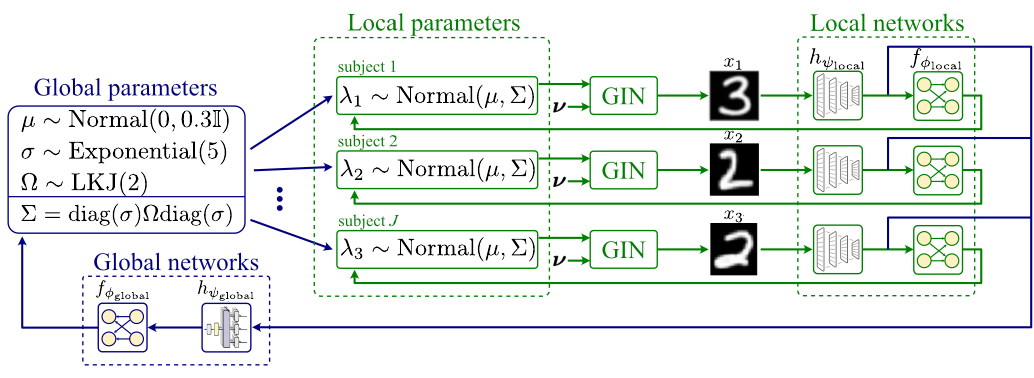}
    \caption{\textbf{Experiment 3.} Overview of the hierarchical model of digit style inference.
    }
    \label{fig:mlm-gin-overview}
\end{figure}

As a final experiment, we demonstrate the efficacy of our amortized multilevel approximator on unstructured, high-dimensional observations. Concretely, the observations are images that are generated from a pre-trained generative neural network which we treat as a black-box simulator (i.e. an implicit data model), see \autoref{fig:mlm-gin-overview} for an overview.

The simulator is a so-called general incompressible flow network \citep[GIN,][]{Sorrenson2020GIN}, which has been trained on a dataset consisting of handwritten digits that are represented as labeled gray-scale images with a dimension of $28\times28$ pixels. GINs have a special architecture that allows them to recover latent dimensions from some observed data $x$, given that the data are also annotated with some conditions $u$ (e.g., class labels or time indices of a time series). The main advantage of GINs over traditional methods for identifying latent variables from observed data such as independent component analysis \citep{comon1994independent} is that the relationship between latent variables and data can be non-linear, while the method is still theoretically well justified. Similarly to the normalizing flows introduced in \autoref{sec:npe}, GINs learn an invertible mapping $g$ from the latent variables $w \in \mathbb{R}^{d}$ to the data space $x = g(w; \phi)$, where g is parameterized by network weights $\phi$.
In their experiment, \citet{Sorrenson2020GIN} trained a GIN on \num{240000} images of handwritten digits and corresponding labels from the EMNIST dataset \citep{cohen2017emnist}. They identified 22 broadly interpretable latent space variables, of which the first 8 variables encode general style features such as slant and line thickness, while the remaining 14 features encode styles of individual digits (e.g., writing a seven with or without a middle stroke).

Because the learned mapping between latent style features and observations is invertible, we can use the trained GIN from \citet{Sorrenson2020GIN} to simulate images conditioned on the style parameters. For this experiment, we restrict ourselves to the first 4 general style parameters (width of top half of the digit, slant, height, and bend through center), treating all other style parameters as random noise. Our multilevel setting features ten subjects $j=1,\dots,10$. Each subject has a 4-dimension style vector $\lambda_j=(\lambda_{j,1}, \lambda_{j,2}, \lambda_{j,3}, \lambda_{j,4})$ which constitute the local parameters $\{\lambda_j\}_{j=1}^{J}$. The global parameters are an average style $\mu \in \mathbb{R}^{4}$, a style dispersion $\sigma \in \mathbb{R}^{4}_{>0}$, and the 6 unique elements of a correlation matrix $\Omega \in [-1,1]^{4 \times 4}$ between styles. The full multilevel model is given by:
\begin{equation}
    \begin{aligned}
        \mu\ &\sim \mathrm{Normal}(0, 0.3\cdot \mathbb{I}),\quad \sigma \sim \mathrm{Exponential}(5), \quad \Omega \sim \mathrm{LKJ}(2),& \hspace*{-10mm}\text{(global parameters)} \\
        \lambda_j &\sim \mathrm{Normal}(\mu, \Sigma)\quad\text{with}\;\Sigma=\mathrm{diag}(\sigma)\,\Omega\,\mathrm{diag}(\sigma)\;\text{and}\;j=1,\ldots,J, & \text{(local parameters)}\\
        x_j &= \mathrm{GIN}(\theta_j, \epsilon)\;\text{with}\;j=1,\ldots,J, & \text{(data model)}
    \end{aligned}
\end{equation}
where $\mathrm{Normal}(\mu, \Sigma)$ is the multivariate normal distribution with mean vector $\mu$ and covariance $\Sigma$, $\mathbb{I}$ is the $4 \times 4$ identity matrix, $\mathrm{Exponential}(5)$ is the exponential distribution with rate 5 , $\mathrm{LKJ}(2)$ is the Lewandowski-Kurowicka-Joe distribution \citep{lewandowski2009} with shape 2, and $\mathrm{GIN}: (\lambda_j,\epsilon) \mapsto x_j$ is the forward pass of the pre-trained global incompressible-flow network from \citet{Sorrenson2020GIN} with style vector $\lambda_j$ and unmodeled style  parameters $\epsilon \sim \mathrm{Normal}(\mu, 0.5)$.

\paragraph{Model training}
The local summary network $h_{\psi_{\text{local}}}$ transforms $28 \times 28$ images to a summary vector of dimension $32$ through a series of residual convolutional blocks followed by average pooling. The global summary network $h_{\psi_{\text{global}}}$ is a set transformer \citep{lee2019set} which learns a 32-dimensional summary vector from the local summary vectors of all subjects. \autoref{sec:supp:training-details} contains details about the neural networks and training settings.

\begin{figure}
    \centering
    \begin{subfigure}[t]{\linewidth}
        \includegraphics[trim={0 6mm 0 14mm},clip,width=0.56\linewidth,valign=c]{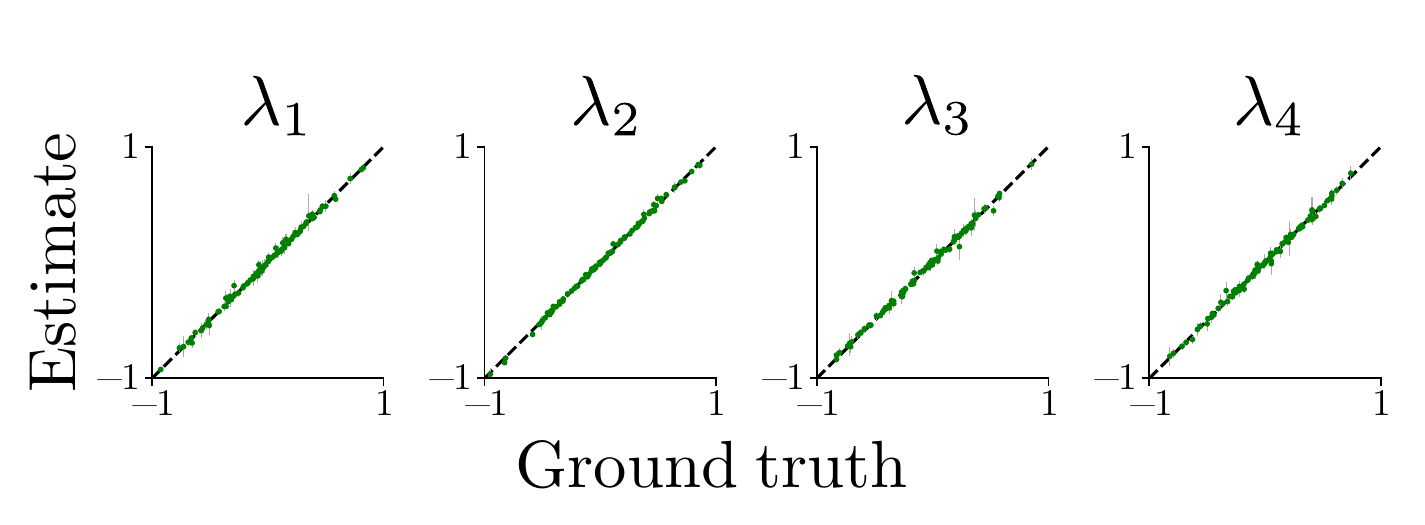}
        \hfill
        \includegraphics[width=0.17\linewidth,valign=c]{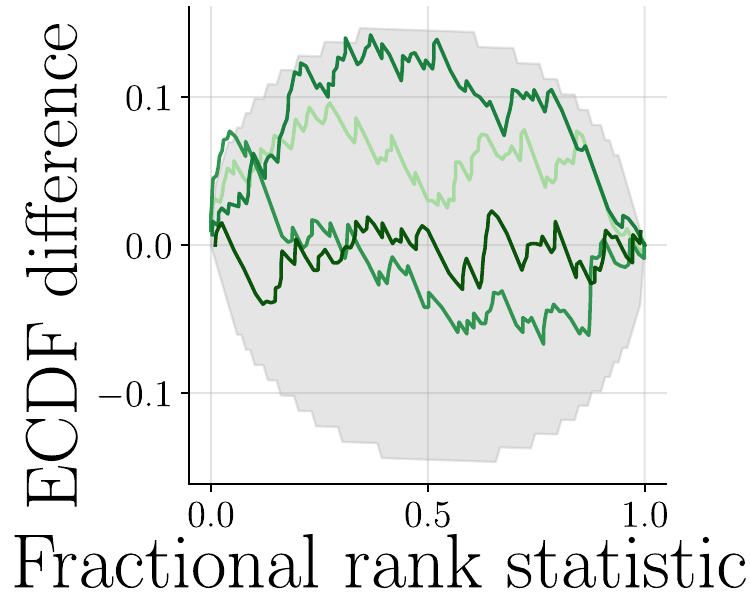}
        \caption{Local parameters of ``subject'' $j=1$ (plots for the remaining subjects are shown in \autoref{sec:supp:additional-results}).}
        \label{fig:mlm-gin-results-local}
    \end{subfigure}
    \begin{subfigure}[t]{\linewidth}
        \includegraphics[trim={0 5mm 0 13mm},clip, width=0.56\linewidth,valign=c]{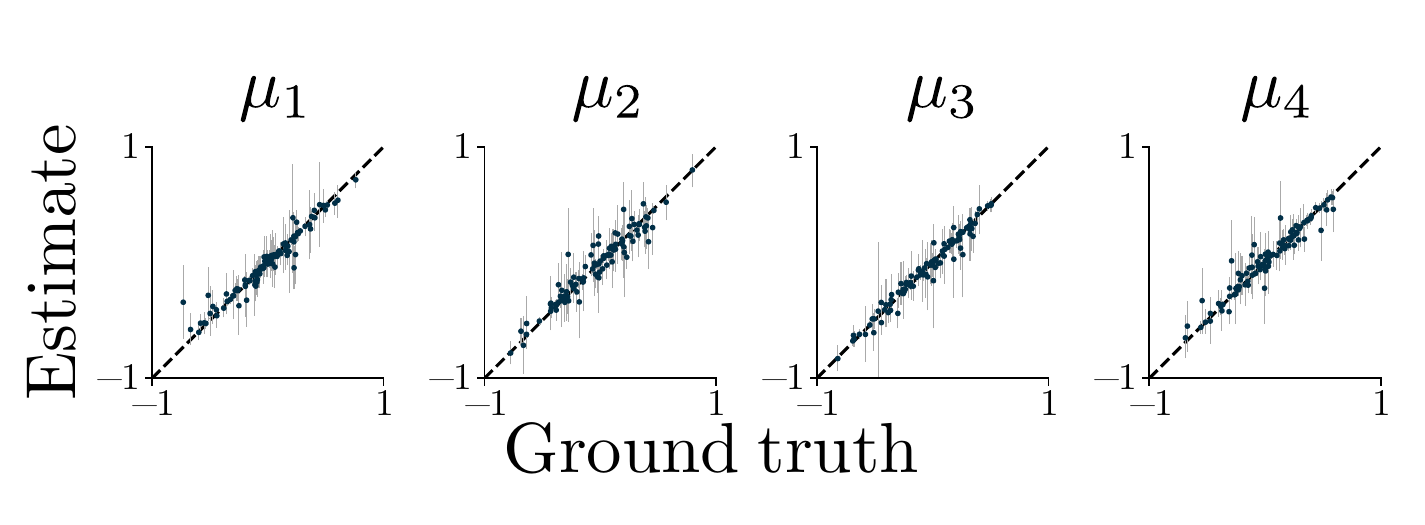}
        \hfill
        \includegraphics[width=0.17\linewidth,valign=c]{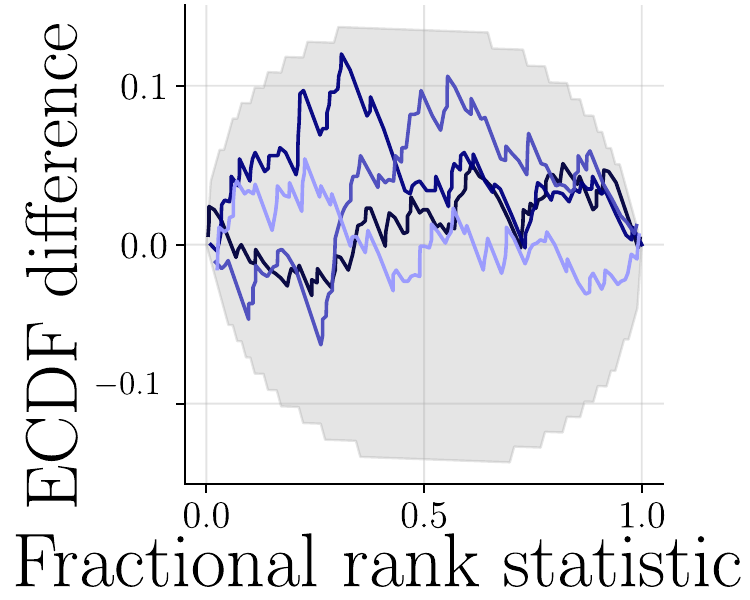}
    \\
        \includegraphics[trim={0 5mm 0 15mm},clip,width=0.56\linewidth,valign=c]{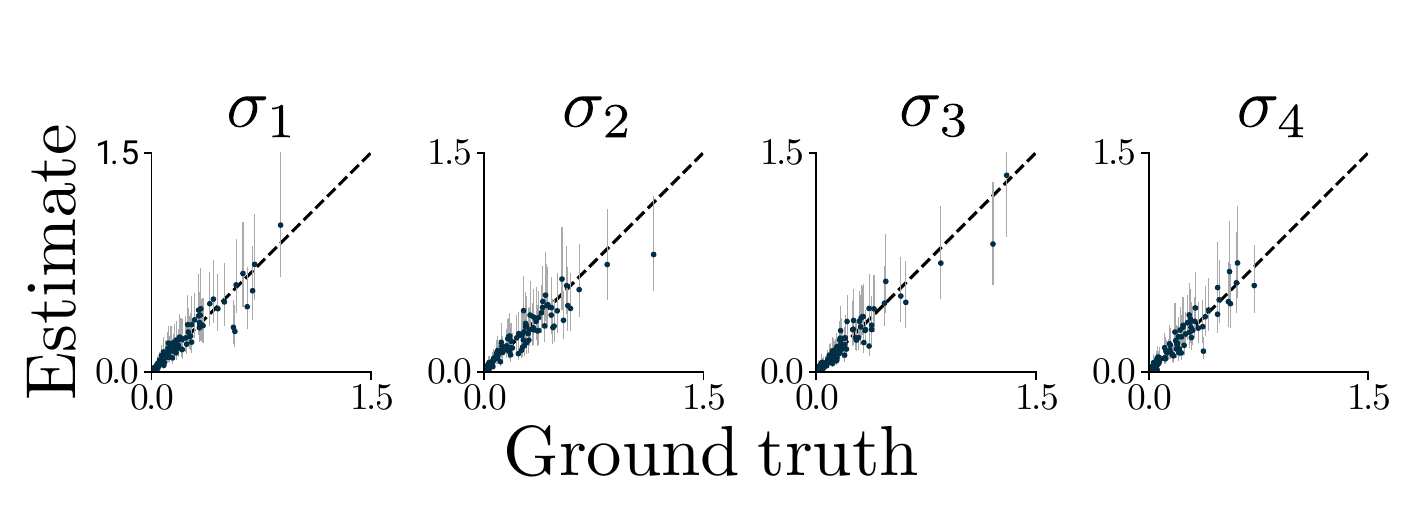}
        \hfill
        \includegraphics[width=0.17\linewidth,valign=c]{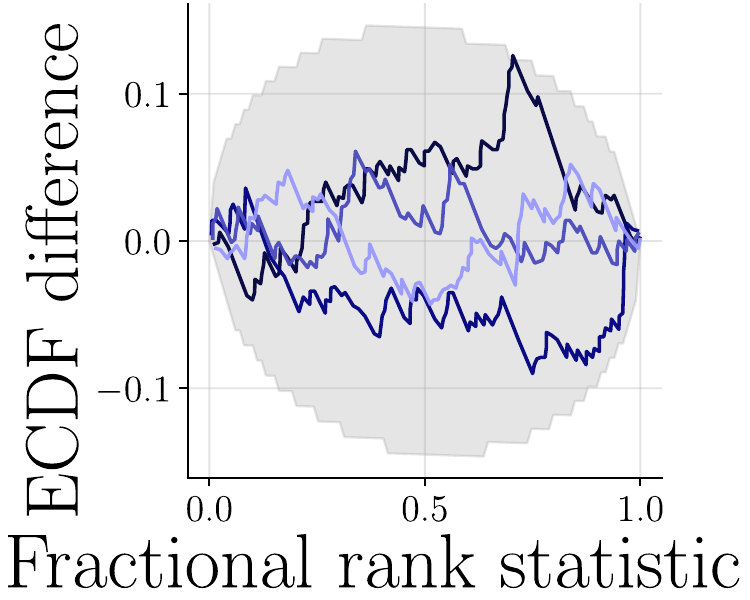}
    \\
        \includegraphics[trim={0 5mm 0 13mm},clip,width=0.82\linewidth,valign=c]{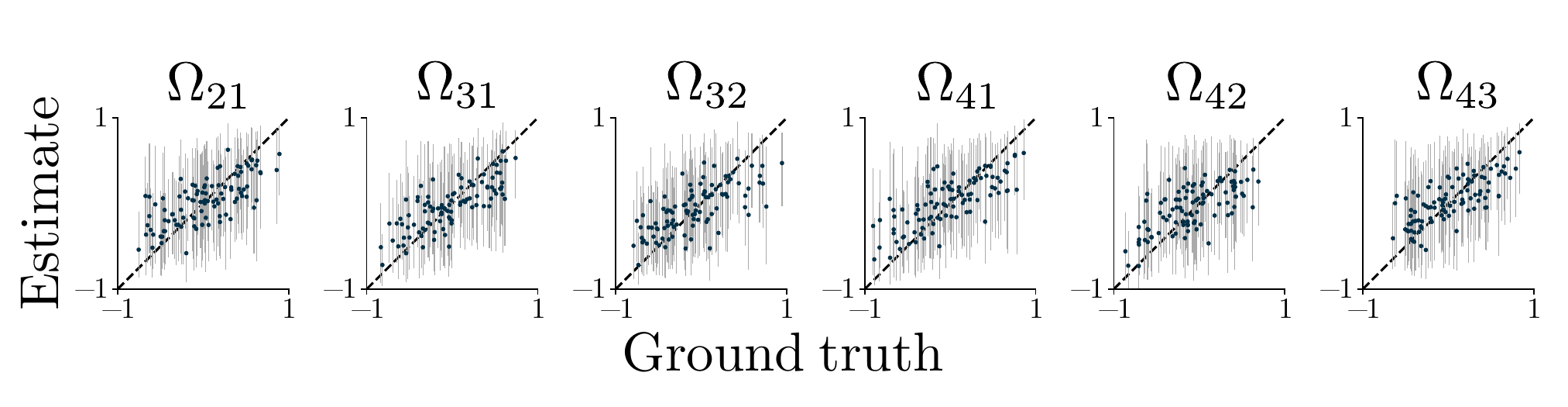}
        \hfill
        \includegraphics[width=0.17\linewidth,valign=c]{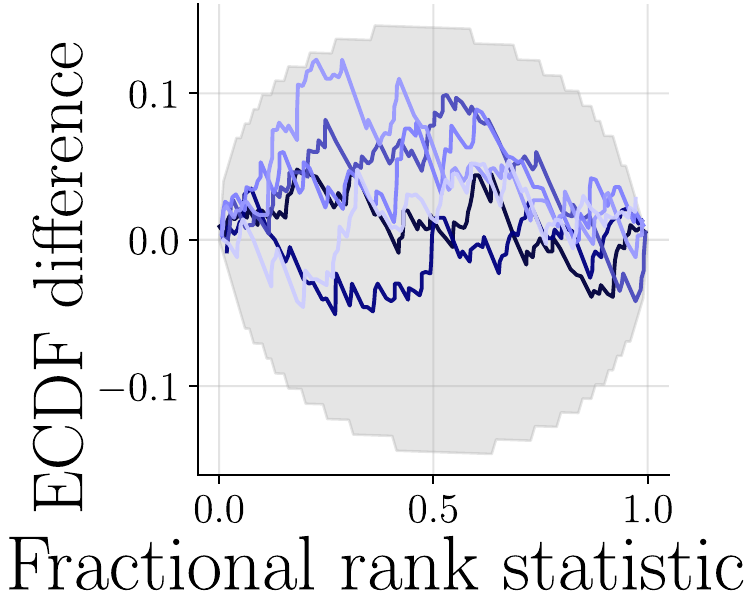}
    \caption{Recovery and calibration of global parameters.}
    \label{fig:mlm-gin-results-global}
    \end{subfigure}
    \caption{\textbf{Experiment 3.} 
    The recovery of the local parameters (\subref{fig:mlm-gin-results-local}, left) and global parameters (\subref{fig:mlm-gin-results-global}, left) is at the upper limit given the epistemic and aleatoric uncertainty in the probabilistic model.
    Dots indicate posterior means, vertical bars represent the symmetric 95\% posterior credible intervals.
    All local and global parameters are well-calibrated (right, SBC-ECDF plots).
    }
\end{figure}

\paragraph{Results}
We draw $1000$ approximate posterior samples from our multilevel neural posterior estimator conditioned on simulated data of $J=10$ subjects and compare posterior estimates to simulated ground truths. We also report posterior calibration of local and global parameters. The estimates of the global parameters $\mu$, $\sigma$, $\Omega$ show excellent calibration (see \autoref{fig:mlm-gin-results-global}). While the average styles $\mu$ and dispersion $\sigma$ can be recovered with near-perfect accuracy, the estimates of $\Omega$ show much greater uncertainty due to the limited sample size of $J=10$ subjects. All local parameters of all subjects can be recovered with good precision and calibration (see \autoref{fig:mlm-gin-results-local}).

\section{Conclusion}
In this work, we developed a framework for amortized Bayesian inference of multilevel models using deep generative neural networks. By utilizing the dependency structure of multilevel models and mirroring them in the neural architectures, we show efficient training of the neural approximators and flexible amortization over both the number of groups and the number of observations within groups. This means that, once trained, the neural approximators provide near instant inference of multilevel models on any number of new datasets or data subsets. We demonstrated amortized Bayesian inference on multilevel models in three realistic experiments, where we show both the amortization capabilities and the accuracy of the resulting Bayesian inference as validated by simulation-based calibration and comparison with Stan. To foster practical application, we implement our methods and architectures in the Python library BayesFlow \citep{radev2023}, which provides efficient and user-friendly workflows for amortized Bayesian inference.

\subsection{Future work}
So far, research on amortized multilevel modeling has mainly focused on two-level models. This already covers a lot of use cases, but still leaves out many practically relevant scenarios of inference on datasets with more than two levels, which should be considered in future work. Another challenge is presented by models with very expensive simulators, such that only very few (say, a maximum of a couple of hundred) simulations are available for training the neural approximators. In this low data setting, fully amortized methods may not achieve acceptable levels of inference accuracy and calibration \citep{lueckmann2021benchmarking,schmitt_self-consistency_2024,radev2023jana,geffner2023compositional}. Instead, sequential neural methods, which enable targeted inference only for a single dataset \citep{papamakarios_fast_2016, greenberg_automatic_2019} may be better suited but have yet to be developed and validated for multilevel models.

\section*{Acknowledgements}
Daniel Habermann, Lars Kühmichel, Stefan Radev, and Paul Bürkner acknowledge support of the Deutsche Forschungsgemeinschaft (DFG, German Research Foundation) Projects 508399956 and 528702768. Paul Bürkner acknowledges
support of DFG Collaborative Research Center 391 (Spatio-Temporal Statistics for the Transition of Energy and Transport) – 520388526. Marvin Schmitt and Paul Bürkner acknowledge support of Cyber Valley Project CyVy-RF- 2021-16, the DFG under Germany’s Excellence Strategy – EXC-2075 - 390740016 (the Stuttgart Cluster of Excellence SimTech). Marvin Schmitt acknowledges travel support from the European Union’s Horizon 2020 research and innovation programme under grant agreements No 951847 (ELISE) and No 101070617 (ELSA), and the Aalto Science-IT project.

\clearpage
\bibliographystyle{ba}
\bibliography{bibliography}

\clearpage

\appendix

\section{Terminology} \label{sec:supp:terminology}
\subsection{Simulation-based inference}

In scientific modeling, the likelihood $p(\theta \given x)$ of a model may often be analytically intractable, for example, because the model contains differential equations without known analytical solutions or several levels of latent variables following complex generative procedures \citep{diggle1984monte, Cranmer2020}.

The Bayesian generative model is then only available as a triple of the prior $p(\theta)$ over simulation parameters $\theta$, a probabilistic model $p(\zeta \given \theta)$ for nuisance parameters (noise) $\zeta$, and a simulation program $g:(\theta, \zeta) \mapsto x$ that outputs synthetic observable data $x$.
More concretely, the forward process is defined as
\begin{equation}\label{eq:sbi-simulation}
    \begin{aligned}
        x=g(\theta, \zeta)\quad\text{with}\quad
        \zeta \sim p(\zeta \given \theta),\;\;
        \theta \sim p(\theta).
    \end{aligned}
\end{equation}
We can readily sample from this \emph{generative model} and obtain pairs of observable data $x$ along with the data-generating parameters $\theta$.
When estimating models with analytically intractable likelihoods, one commonly speaks of \textit{likelihood-free inference}, a term that is somewhat misleading as the likelihood still \emph{implicitly} exists as an integral over all possible execution paths (as represented by the stochastic noise variates $\zeta$),
\begin{equation}
    p(x \given \theta) = \int p(x,\zeta \given \theta)\mathrm{d}\zeta,
\end{equation}
even though the analytic form of that likelihood $p(x \given \theta)$ is generally unknown \citep[][]{Cranmer2020}.
Nevertheless, Monte Carlo simulations from the full model (\autoref{eq:sbi-simulation}) can be used to perform fully Bayesian inference, which motivates the alternative and more descriptive term \textit{simulation-based inference} (SBI). 
Intractable models that require SBI are ubiquitous in the quantitative sciences and SBI has been conceptualized as one of the key pillars of today's computational stack \citep{lavin2021simulation}.

Simulation-based methods are also applicable when the likelihood is tractable, as is the case in many MLM applications. 
Traditionally, for tractable likelihood models, simulation-based methods were not considered because they were inferior to density-based inference methods (e.g., MCMC) in terms of both inference speed and accuracy. 
However, simulation-based training of neural networks seems crucial for achieving \textit{amortized Bayesian inference} via neural networks, regardless of whether we have access to the analytic likelihood (or prior) densities.

\subsection{Amortized Bayesian Inference}
While performing Bayesian inference for high-dimensional models on a single data set is already challenging, the computational burden increases considerably when the same model must be applied to multiple data sets. 
Achieving accurate posterior inference can quickly become too computationally expensive in such cases. This issue is most apparent when fitting the same statistical model to several or sometimes millions \citep{Krause2022} of mutually independent data sets.

Amortized Bayesian Inference (ABI) \citep[ABI;][]{gershman2014amortized, ritchie2016deep, le2017inference, Radev2020} offers a promising solution to the problem of repeated refits for models with or without tractable likelihood functions. 
ABI decouples estimation into two phases: (1) a training phase, during which a neural network learns a posterior functional from model simulations, and (2) an inference phase, where posterior sampling is transformed into an almost instantaneous forward pass through the neural network.
The term \textit{amortized} emphasizes that the computational demand from the initial training phase amortizes over subsequent rapid model fits, typically within a fraction of a second when using invertible neural networks \citep{kruse2021benchmarking}.

Towards amortized Bayesian inference, several methods have been developed that amortize the evaluation of certain model or algorithm components in combination with other non-amortized components. Examples include amortized approximate ratio estimators \citep{hermans2020likelihood} and the likelihood approximation networks of \citet{Fengler2021} which can be used within (non-amortized) MCMC, as well as amortized methods to improve proposal distributions within (non-amortized) sequential Monte-Carlo.

\subsection{Neural Density Estimation}

Neural density estimation (NDE) approximates a target distribution $p(x)$ with a surrogate density $q_{\phib}(x)$, where $\phib$ denotes learnable weights of a generative neural network $f_{\phib}$. 
This so-called \textit{inference network} $f_{\phib}$ often implements a normalizing flow \citep{kobyzev2020normalizing} through a conditional invertible neural network \citep{ardizzone2019guided} capable of modeling complex target distributions. 
Such networks have demonstrated remarkable success in tackling amortized neural posterior estimation (NPE) problems across disciplines \citep[e.g.,][]{ardizzone2018analyzing, gonccalves2020training, Bieringer2021, Krause2022, avecilla2022neural}, where the target distribution is the posterior.
That said, other generative neural architectures, such as masked autoregressive flows \citep{papamakarios2017maf} can be used for NDE as well and are fully compatible with our proposed methods.

\section{Alternative factorization} \label{sec:supp:alternative-factorization}

The factorization of the posterior of a two-level model shown in the main manuscript is not unique. An equally valid factorization is:
\begin{equation}
    \begin{split}
        p(\tau ,\omega, \{\lambda_j\} \given x) &=  p(\{\lambda_j\} \given x)p(\tau \given \{\lambda_{j}\})\,p(\omega \given x, \{\lambda_j\}),
\end{split}
\end{equation}
which suggests a different neural inference architecture consisting of three components
$f_{\phi_1}(\{\lambda_j\};h_{\psi_1}(\{x_j\}))$,
$f_{\phi_2}(\tau; h_{\psi_2}(\{\lambda_{j}\}))$ and 
$f_{\phi_3}(\omega; ,h_{\psi_{3A}}(\{\lambda_j\}),h_{\psi_{3B}}(\{x_j\}))$,
where the notation $\{\lambda_j\}$ represents the set of all local parameters across all groups, while $\{x_j\}$ denotes the complete dataset, encompassing the data from all groups $j$.

For this factorization, we utilize the fact that the hyper parameters $\tau$ are completely identified by the local parameters $\lambda_j$. We now also require an additional inference network solely for the shared parameters $\omega$, as they are not identified by the local parameters alone. 
Compared to the factorization in the main manuscript, this architecture cannot easily be amortized over the number of groups, because we lack per-group summary inference networks (i.e., the inference network $f_{\phi_1}(\{\lambda_j\};h_{\psi_1}(\{x_j\}))$ has to estimate all local parameters jointly).

Amortization capabilities can be restored by modifying the inference network $f_{\phi_1}$ so it can estimate each group independently. This could be achieved by splitting the single summary network $h_{\psi_1}$ into two modules, one that provides a global representation of the global structure (i.e, sufficient statistics to determine the degree of required shrinkage), and one that learns representation of the local data structure of group $j$: 
\begin{equation}
f_{\phi_1}(\lambda_j;h_{\psi_{1A}}(\{x_j\}), h_{\psi_{1B}}(x_j)).
\end{equation}
This restores amortization capabilities and keeps the output dimensionality of all inference networks constant regardless of the dataset. However, the implied architecture is more involved, requiring up to 5 summary and 3 inference networks, instead of only 2 summary and 2 inference networks for the factorization presented in the main manuscript.  
While some further optimizations are possible, for example by combining the summary networks $h_{\psi_{2}}$ and $h_{\psi_{3A}}$ as well as $h_{\psi_{1}}$ and $h_{\psi_{3B}}$, we decided to further study only the structurally simpler factorization presented in the main manuscript in our empirical evaluations.

\section{In which situations should (ML-)NPE be considered over traditional MCMC sampling?} \label{sec:supp:NPE-over-MCMC}

Our current recommendation with regards to the use of amortized inference methods vs. HMC is the following: If a problem can be adequately addressed using HMC, then it should be. In our point of view, the field of amortized Bayesian inference using NPE is exciting and even has the potential to replace HMC as the go-to method for most inference tasks, but it is still a developing field. In particular, the strong diagnostics and reliability of Stan and other tools are unmatched. If a user runs a Stan model and it returns no warnings, they can be very sure that they retrieved an accurate representation of the typical set; it mostly just works.
This experience is, of course, the result of a great amount of work by the PyMC3/Stan/Turing community and others.
However, we believe that NPE methods have reached a level of maturity that already makes them interesting for many applications. This is particularly the case if:

\begin{itemize}
    \item a model needs to be repeatedly fit on new data
    \item model fits need to be exceptionally quick (e.g., real time applications)
    \item a principled Bayesian workflow with cross-validation and SBC is desired and a single model run is already expensive
    \item a model has hierarchical structure and there are many groups, such that the model becomes too slow to run using MCMC sampling
    \item a model requires solving partial differential equations or the likelihood is expensive to evaluate for other reasons
\end{itemize}

Depending on the inference task, NPE can already be multiple orders of magnitude quicker, even if one is only interested in evaluating the posterior for a single data set without SBI or cross-validation.
While many models can be fit using NPE on a standard desktop computer, NPE is also well-positioned to take advantage of high-performing GPUs and compute clusters and scales more favorably to larger problems (e.g., in terms of the number of model parameters) than HMC.

\section{Network Training Details} \label{sec:supp:training-details}

\subsection{Experiment 1: Air passenger traffic}
To amortize over the number of countries and observed time points, we vary the number of groups per dataset $J_m \sim \text{DiscreteUniform}(10, 30)$ and the number of time points $T_m \sim \text{DiscreteUniform}(5, 30)$ for each training sample $m$.

The local summary network has an initial long short-term memory \citep[LSTM;][]{Hochreiter1997} layer with 64 output units to capture time dependencies at the group level, followed by a dense layer with 256 units and exponential linear unit activation function and another dense layer with 32 units. For the global summary network, we employ a set transformer \citep{Lee2018} with 32 summary dimensions, 16 inducing points and 2 hidden layers with 64 units with a rectified linear unit \citep[ReLU][]{Agarap2018} activation function. The multihead attention layer of the set transformer uses an attention block with 4 heads and key dimension 32 with 1\% dropout. The inference networks are implemented as neural spline flows \citep{Durkan2019} with 6 coupling layers using 2 dense layers with 256 units each, a ReLU activation function and 5\% dropout. We set the simulation budget to $M=100000$ and train for 200 epochs using the Adam optimizer \citep{Kingma2014} with an initial learning rate of $5\times10^{-4}$ with cosine decay \citep{Loshchilov2016} and a batch size of 32.

\subsection{Experiment 2: Diffusion Decision Model}
To amortize over the number of subjects, we vary the number of subjects per dataset $J_m \sim \text{DiscreteUniform}(10, 30)$ and the number of observations per subject $N_m \sim \text{DiscreteUniform}(1, 100)$.
The local summary network consists of a set transformer with 16 summary dimensions, 32 inducing points, 2 hidden layers with 64 units with a ReLU activation function and Glorot uniform initialization \citep{Glorot2010}. The multihead attention layer utilizes an attention block with 4 heads and key dimension 32 with 1\% dropout. The architecture of the global summary network is identical to the local summary network, except using 16 inducing points and 32 summary dimensions. For the inference networks, we use neural spline flows with 4 hidden layers for both local and global approximators. We set the simulation budget to $M=20000$ and train for 200 epochs using the Adam optimizer \citep{Kingma2014}  with an initial learning rate of $5\times10{-4}$ with cosine decay \citep{Loshchilov2016} and a batch size of 32.

\subsection{Experiment 3: Style inference for hand-drawn digits}
For this experiment, we don't amortize over the number of subjects per dataset and fix $J_m = 10$.
The global inference network is an affine coupling flow with 6 coupling layers, each of which features 2 dense layers with 128 units, $5\%$ dropout, and kernel regularization with strength $10^{-4}$. The local inference network is an affine coupling flow with 10 coupling layers, each of which features one dense layer with 512 units, no dropout, and kernel regularization with strength $10^{-4}$.
The local summary network is a convolutional neural network (CNN) with residual connections, Mish activation \citep{misra2019mish}, and He Normal initialization \citep{he2015delving}, followed by average pooling. The filter sizes of the convolutional layers are 32--64--32--32--32 (output).
The global summary network is a set transformer with 32 output dimensions, 32 inducing points, 2 dense layers, and 2 attention blocks. The neural networks are trained for 200 epochs with a batch size of 128 and 500 iterations per epoch.
We use the Adam optimizer \citep{Kingma2014} with an initial learning rate of $10^{-3}$ and cosine decay \citep{Loshchilov2016}.

\newpage
\section*{\texorpdfstring{Supplementary D: Additional results}{Additional results}} \label{sec:supp:additional-results}
\begin{figure}[H]
    \centering
    \includegraphics[width=0.6\linewidth]{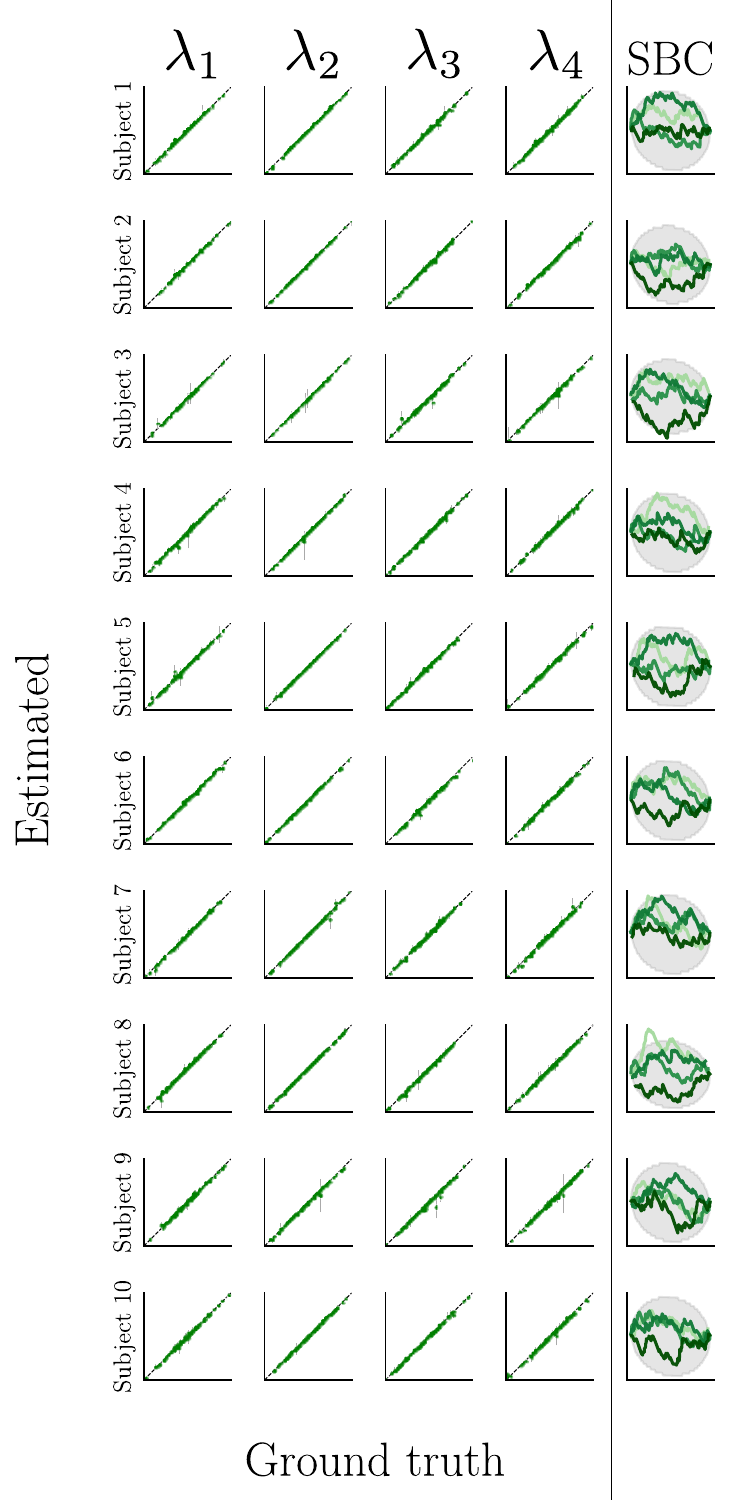}
    \caption{\textbf{Experiment 3.} Posterior recovery and simulation-based calibration for all 10 subjects.}
    \label{fig:mlm-gin-recovery-calibration-local-all}
\end{figure}

\end{document}

%% file: _math.tex
\usepackage{natbib}
\usepackage{amsmath}

\newcommand{\phib}{\boldsymbol{\phi}}

\newcommand{\given}{\,|\,}

\DeclareMathOperator*{\argmin}{arg\,min}